\documentclass[11pt]{article}

\usepackage[preprint]{acl}

\usepackage{times}
\usepackage{latexsym}

\usepackage[T1]{fontenc}

\usepackage[utf8]{inputenc}

\usepackage{microtype}

\usepackage{inconsolata}

\usepackage{graphicx}

\usepackage[utf8]{inputenc} 
\usepackage[T1]{fontenc}    
\usepackage{hyperref}       
\usepackage{url}
\usepackage{booktabs}
\usepackage{multirow}
\usepackage{makecell}
\usepackage[toc,page]{appendix}
\usepackage{titletoc}
\usepackage{amsfonts}       
\usepackage{nicefrac}       
\usepackage{microtype}      
\usepackage[normalem]{ulem}
\useunder{\uline}{\ul}{}
\usepackage[dvipsnames]{xcolor}
\usepackage[table,dvipsnames]{xcolor}
\definecolor{rubricblue}{RGB}{238,244,250}
\definecolor{rubricgray}{RGB}{247,247,247}
\definecolor{rubricred}{RGB}{252,239,239}

\newcommand{\bench}{\textsc{Curation-Bench}}

\usepackage{amssymb}

\usepackage{algorithm}
\usepackage{algpseudocode}
\usepackage{wrapfig}
\usepackage{lipsum}
\usepackage{subcaption}
\usepackage[dvipsnames]{xcolor}
\usepackage{microtype}
\usepackage{graphicx}
\usepackage{subcaption}
\usepackage[hashEnumerators, smartEllipses]{markdown}
\usepackage{longtable}
\usepackage{array}
\usepackage{listingsutf8}
\usepackage{upquote}
\usepackage{subcaption}
\usepackage{mdframed}

\usepackage{tikz}
\usetikzlibrary{positioning, fit, backgrounds, arrows.meta}

\usepackage{pifont}

\newcommand{\dataset}{\mathcal{D}}
\newcommand{\policy}{\pi}

\newcommand{\yes}{\ding{51}}
\newcommand{\no}{\ding{55}}

\usepackage{enumitem}
\newcolumntype{C}[1]{>{\centering\arraybackslash}p{#1}}
\usepackage{lineno}

\definecolor{darkblue}{rgb}{0, 0, 0.5}
\hypersetup{colorlinks=true, citecolor=darkblue, linkcolor=darkblue, urlcolor=darkblue}

\definecolor{rubricHeader}{RGB}{232,240,248}
\definecolor{rubricRowA}{RGB}{250,252,255}
\definecolor{rubricRowB}{RGB}{244,248,252}
\definecolor{baselineRow}{RGB}{246,248,250}
\definecolor{keepRow}{RGB}{244,250,246}
\definecolor{discardRow}{RGB}{250,250,250}
\definecolor{bestRow}{RGB}{232,245,233}
\usepackage{tabularx}
\newcolumntype{Y}{>{\raggedright\arraybackslash}X}

\definecolor{scafBase}{RGB}{247,247,247}
\definecolor{scafLight}{RGB}{241,247,255}
\definecolor{scafHeavy}{RGB}{245,250,244}
\definecolor{scafHeader}{RGB}{232,240,248}

\usepackage{amsmath}
\usepackage{amssymb}
\usepackage{mathtools}
\usepackage{amsthm}

\usepackage[T1]{fontenc}    
\usepackage{amsfonts}       
\usepackage{nicefrac}       
\usepackage{microtype}      %
\usepackage[most]{tcolorbox}
\usepackage{enumitem}
\usepackage{listings}

\usepackage[capitalize,noabbrev]{cleveref}

\theoremstyle{plain}

\theoremstyle{definition}

\theoremstyle{remark}

\usepackage[textsize=tiny]{todonotes}
\title{Can Generalist Agents Automate Data Curation?}

\author{
 \textbf{Feiyang Kang\textsuperscript{1,*}},
 \textbf{Hanze Li\textsuperscript{1,*}},
 \textbf{Adam Nguyen\textsuperscript{1}},
 \textbf{Mahavir Dabas\textsuperscript{1}},
 \vspace{0.2em}\\
 \textbf{Jiaqi W. Ma\textsuperscript{2}},
 \textbf{Frederic Sala\textsuperscript{3}},
 \textbf{Dawn Song\textsuperscript{4}},
 \textbf{Ruoxi Jia\textsuperscript{1}},
\vspace{0.5em}\\
 \textsuperscript{1}Virginia Tech,
 \textsuperscript{2}University of Illinois Urbana-Champaign,\\
 \textsuperscript{3}University of Wisconsin-Madison,
 \textsuperscript{4}University of California, Berkeley
\\
 \small{
   \textsuperscript{*}Equal Contribution\quad \textbf{Correspondence:} \href{mailto:fyk@vt.edu}{fyk@vt.edu}
 }
}

\begin{document}
\maketitle
\begin{abstract}
Curating training data is among the most consequential yet labor-intensive parts of modern AI development: practitioners iteratively propose, implement, evaluate, and revise data policies against noisy benchmark feedback. We ask whether generalist coding agents can automate this data-curation loop. We introduce \bench{}, an agent-centric benchmark that fixes the model, training recipe, and evaluation suite while giving agents command-line access to inspect data, implement policies, submit them to a fixed training/evaluation pipeline, and revise.
In a vision-language instruction-tuning instantiation, out-of-the-box agents reach strong published data-selection baselines within ten iterations. However, trajectory analysis reveals a persistent \emph{execution--research gap}: agents mainly tune local policy variants rather than explore new policy families, even when given strategy guides and paper references. Scaffolds requiring each iteration to cite, instantiate, and adapt a prior method shift agents toward method-guided exploration. The scaffolded agent autonomously composes---without human design input---a data-selection policy that outperforms strong published baselines at one-tenth their data budget. Overall, current agents can run the curation loop, but reliable data research requires scaffolded method adaptation, not open-ended prompting alone.\footnote{Code and benchmark available at: \texttt{\url{https://github.com/feiyang-k/curation-bench}} .\vspace{-0em}}


\end{abstract}

\section{Introduction}
\begin{figure}[h]
\centering
\includegraphics[width=\linewidth]{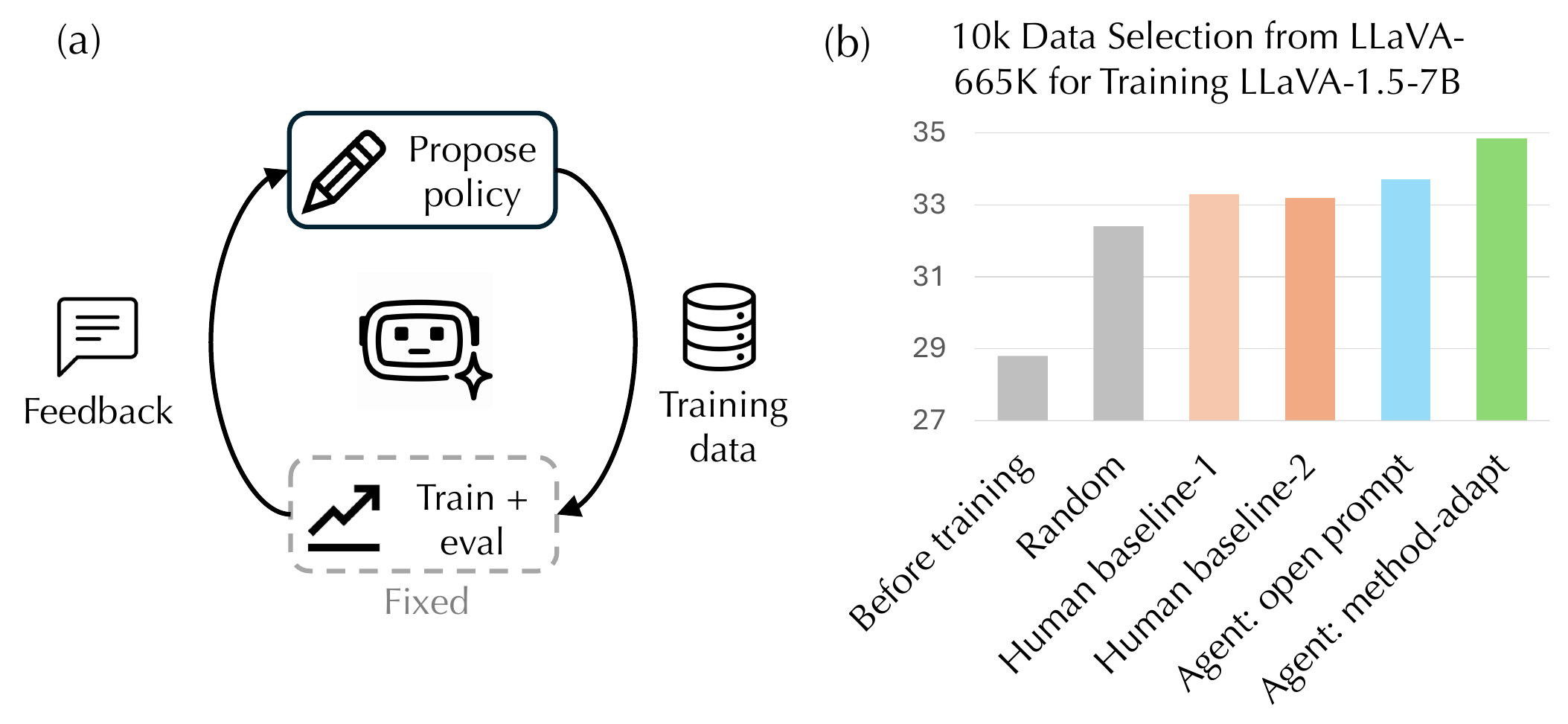}\vspace{-0.5em}
\caption{\small{\textbf{Agentic data curation requires more than open-ended prompting.}
(a) We formulate data curation as a policy-search loop: the agent proposes a data policy, constructs training data, observes feedback from fixed training and evaluation, and revises the policy.
(b) On 10k-example selection from LLaVA-665K dataset for LLaVA-1.5-7B training, open-ended prompting improves over random selection and human-designed baselines~\cite{wu2024icons,yang2025data}, but method-adaptation scaffolding achieves the best result by adapting prior data-selection methods into a stronger policy.
}}\vspace{-1.5em}
\label{fig:intro}
\end{figure}

Training data is now a first-class design variable in modern AI: it shapes what models can do, how they fail, and how efficiently they learn. Prior work has automated many individual data operations, including pruning~\citep{sorscher2022beyond,paul2021deep}, deduplication~\citep{lee2022deduplicating,abbas2023semdedup,kang2025adadedup}, remixing~\citep{xie2023doremi,fan2024doge,chen2026olmix}, rewriting~\citep{wang2024template,maini2024rephrasing}, and synthetic generation~\citep{wang2023self,alpaca,kang2025demystifying}. What remains largely manual is deciding which technique to use, how to adapt it, and when to revise it after noisy benchmark feedback. Because no technique is universally effective~\citep{zhang2025best,goyal2024science}, practitioners must repeatedly propose, implement, evaluate, and refine data policies. This iterative search is where human effort accumulates. We ask whether the search process itself can be automated.

Prior work captures two ingredients needed for automated data curation---controlled data evaluation and iterative experimentation---but not their combination. Data-centric benchmarks such as DataComp~\citep{gadre2023datacomp,li2024datacomp} fix the model, training recipe, and evaluation so that data becomes the variable of interest. This isolation principle is essential, but the submitted artifact is a static policy, and the benchmark does not evaluate how that policy was discovered, revised, or justified. Autonomous ML-research agents~\citep{yang2024swe,lu2024ai,kon2026expbench,karpathy2026autoresearch} evaluate iterative behavior, but typically optimize training recipes---code, hyperparameters, architectures, or model dimensions---under a fixed data distribution~\citep{chan2024mle,karpathy2026autoresearch}. Data-science agents~\citep{pinchuk2026tml,jing2024dsbench} iterate over analysis pipelines and predictive models, treating data as input rather than as the object of optimization. What remains missing is an agentic benchmark for iterative data-policy discovery.

We introduce \bench{} to fill this gap. \bench{} inherits DataComp's data-isolation principle by fixing the model, training recipe, and evaluation suite, but replaces static policy submission with an interactive terminal loop: the agent inspects the candidate pool, implements curation policies, submits them to a fixed training/evaluation pipeline, observes feedback, and revises. \bench{} records the full curation trajectory---policy scripts, data manifests, training outputs, and evaluation logs---so that agents can be evaluated not only by final accuracy, but also by how they search: whether they explore new data-policy families, ground decisions in prior evidence, or fall back on shallow local adjustments, and whether those choices improve performance.

We study the most immediate candidates for automating iterative data-policy discovery: generalist coding agents, off-the-shelf terminal systems not designed specifically for data curation. They can already inspect files, write scripts, launch jobs, parse logs, and recover from failures. We ask how far such generic execution capabilities can go in data curation, and what additional guidance helps agents make better data-policy decisions.

We instantiate \bench{} primarily in multimodal instruction tuning, where candidate pools mix heterogeneous data sources, evaluations cover diverse visual and linguistic skills, and no single selection heuristic is uniformly effective; we further include a smaller-scale CLIP-style pretraining setting to test whether the framework extends beyond instruction tuning alone. The main instantiation yields two findings. First, generalist agents are already useful data-curation executors: under open-ended prompting, they reach strong published data-selection baselines within ten iterations, recovering roughly 60\% of the full-data fine-tuning gain at 1.5\% of the original pool. Second, trajectory diagnostics reveal a persistent execution--research gap: agents reliably run the loop but mainly tune local policy variants---source ratios, length thresholds, and random seeds---rather than explore distinct policy families. This persists even with strategy guides and paper references, suggesting that the bottleneck is not missing methodological knowledge but failure to operationalize it into executable data policies. We therefore study scaffolds as interventions and find that mandatory citation, instantiation, and adaptation of prior methods shift agents toward method-guided exploration of new policy families. Under the strongest scaffold, the agent composes an EL2N-style top-loss policy~\cite{paul2021deep} with an assistant-loss noise filter, producing a hybrid loss-based data-selection policy that outperforms the compared published non-agent baselines using one-tenth as much data.

Finally, longer-session experiments suggest that agent search iterations are a
meaningful form of curation compute. When we increase the agent budget beyond
ten iterations, average outcomes continue to improve up to 50 iterations without
a clear plateau. This connects agentic data curation to a broader question of
how to improve performance in a \emph{finite-data, increasing-compute} regime:
when additional raw data is unavailable or increasingly costly, extra compute
can be spent not only on reusing data during training or expanding it
synthetically~\citep{muennighoff2023scaling,kim2026data}, but also on searching
over how to select, adapt, and validate the data already available.

\section{\bench{}}
\label{sec:bench}


This section specifies the benchmark contract used throughout the paper. 
A \bench{} task presents an agent with a candidate data pool and a fixed 
training--evaluation pipeline, and asks the agent to produce a budgeted 
training set through an executable data policy. The benchmark constrains 
what the agent may change, validates each submitted dataset before training, 
and records the full sequence of policy attempts and outcomes. This contract 
separates two evaluation targets: the quality of the final curated data and 
the quality of the agent's search process.
We first describe the design principles that define this contract, then give 
the task formalization, environment interface, and scoring metrics.

\vspace{-0.5em}
\paragraph{Design principles.}
The benchmark contract follows four principles.
\textbf{(P1) Data isolation.}
The model architecture, optimizer, training schedule, and evaluation suite are fixed by the harness. The agent's only controlled variable is the curated data.
\textbf{(P2) Terminal realism.}
The agent operates through a standard terminal workspace, mirroring how current generalist coding agents interact with repositories: inspecting files, writing scripts, running commands, debugging failures, and reading logs.
\textbf{(P3) Contamination control.}
Every submitted dataset must pass an automatic audit before training. The audit checks exact matches and high-overlap text spans against evaluation question--answer pairs, preventing direct leakage from becoming a trivial path to improvement.
\textbf{(P4) Trajectory legibility.}
Each iteration persists the curation script, generated manifest, audit result,
training output, evaluation scores, and run notes under a commit hash. These
records support the trajectory diagnostics in \Cref{sec:bench_diagnostics}.

\begin{figure}[t!]
\centering
\includegraphics[width=\linewidth]{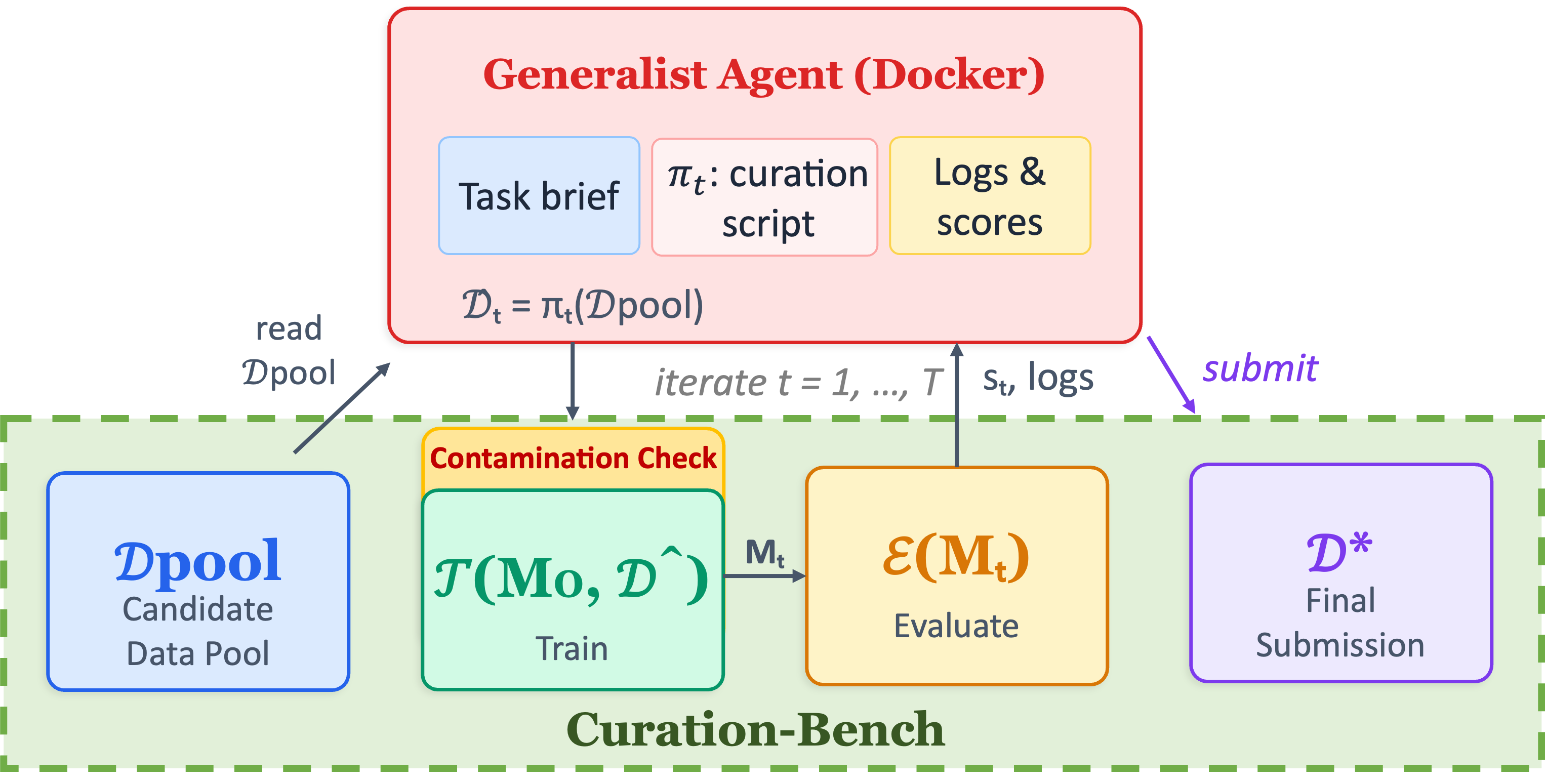}
\caption{\small{\textbf{Architecture of \bench{}}. A coding agent inspects the candidate pool, implements data
policies, and submits curated datasets. The harness validates each submission
and scores it with a fixed training--evaluation pipeline. 
}}\vspace{-1.5em}
\label{fig:benchmark-loopx2}
\end{figure}

\begin{figure*}[h]
\centering
\includegraphics[width=\linewidth]{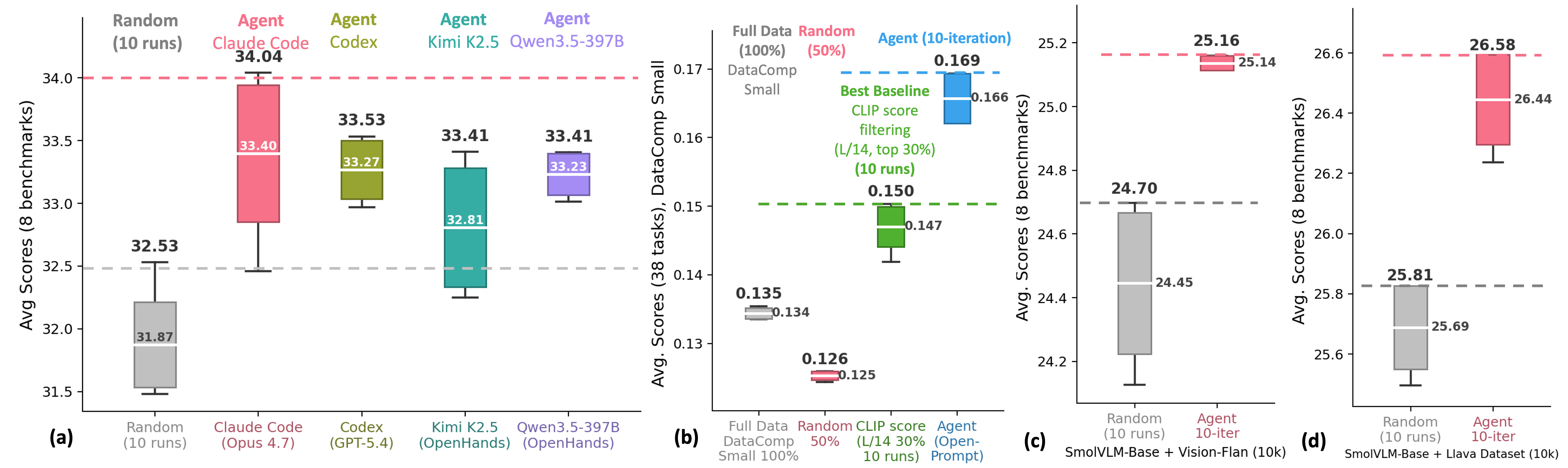}
\vspace{-1.5em}
\caption{\small{
\textbf{Agent data-curation results (open-prompt).} The bars and the centerline show the average scores and the range of standard deviation, and the upper/lower wicks showing the min/max outcome of a session. Agents run 3 sessions each with 10 iterations, and the baseline is repeated for 10 independent runs. 100\% iterations are successfully executed.
\textbf{(a).} Fine-tuning pre-trained LLaVA-1.5-7B on curated subsets of 10k samples from the LLaVA-665K instruction-tuning pool and evaluate on 8 benchmarks. All agents--Claude Code (Opus 4.7), Codex (GPT-5.4), and Kimi K2.5/Qwen3.5-397B with OpenHands harness--significantly improve training outcome compared to training on random data. \textbf{(b).} Pre-training ViT-B/32 CLIP models for the DataComp Small benchmark (Filtering track) and reporting average scores from the 38 tasks. The agent (Claude Code) established a notable margin over the best baseline (top 30\% CLIP L/14 scores). \textbf{(c)}/\textbf{(d).} Fine-tuning SmolVLM-Base on curated subsets of 10k samples from the LLaVA-665K/Vision-Flan 186k instruction-tuning datasets and evaluate on 8 benchmarks, respectively. The agent (Claude Code) substantially improves training outcome compared to training on random data.
}}
\vspace{-1em}
\label{fig:main_4in1}
\end{figure*}
\vspace{-0.5em}
\subsection{Environment and Tasks}
\label{sec:bench_env}

A \bench{} task is defined by
$
\tau = (M_0,\ \dataset_{\mathrm{pool}},\ B,\ \mathcal{T},\ \mathcal{E},\ \mathcal{C}),
$
where $M_0$ is the initial model, $\dataset_{\mathrm{pool}}$ is the candidate
training pool, $B$ is the data budget, $\mathcal{T}$ is the fixed training
procedure, $\mathcal{E}$ is the evaluation suite, and $\mathcal{C}$ specifies
constraints such as schema, budget, and contamination rules. The agent produces
a data policy $\policy$ that maps the pool to a curated dataset
$
\dataset_{\policy} = \policy(\dataset_{\mathrm{pool}}), \quad
|\dataset_{\policy}| \le B,
$
and the resulting policy is scored by
$
S(\policy) = \mathcal{E}\!\left(\mathcal{T}(M_0,\ \dataset_{\policy})\right).
$

The agent operates in a Dockerized terminal workspace containing a task brief,
read-only access to the candidate pool, a starter random-selection script, a
logging directory for prior iterations, and a benchmark CLI. The CLI supports
dataset validation, contamination audit, training, and evaluation. The agent may
edit the curation script and materialize new datasets, but may not modify the
training recipe, evaluation code, or contamination checks.

At iteration $t$, the agent writes or revises a curation script $\policy_t$,
materializes a candidate dataset $\dataset_t$, and submits it through the CLI.
Submission is gated: the harness validates schema and budget, runs the
contamination audit (Appendix~\ref{app:contam}), and only then launches training. Failed submissions are
returned to the agent before training. Accepted submissions are trained from
$M_0$ using $\mathcal{T}$, evaluated by $\mathcal{E}$, and returned to the agent
with aggregate and per-benchmark scores. The agent may then use this feedback to
design $\policy_{t+1}$.

\subsection{Scoring and Trajectory Diagnostics}
\label{sec:bench_diagnostics}

Agent sessions are evaluated along two axes: downstream outcome and trajectory quality.

\vspace{-0.5em}\paragraph{Outcome metrics.}
For a session of $T$ iterations, the primary outcome is the best score achieved during the session. 
We report absolute score, improvement over the base model and over random selection at the same budget, and the fraction of the full-data fine-tuning gain recovered (\%) when possible. The latter metric provides a more straightforward measure of data efficacy.

\vspace{-0.5em}\paragraph{Trajectory diagnostics.} Outcome scores do not fully characterize the quality of agentic data curation: a high score
may come from systematic exploration, but it may also come from repeated local
tuning or lucky variation. We therefore annotate each iteration with four
labels: \emph{new policy family}, \emph{grounded}, \emph{effective}, and
\emph{shallow}, following the rubrics in Table~\ref{tab:rubric}. We report
these labels both at the iteration level, for individual
inspect--curate--train--evaluate loops, and at the trajectory level, for the
full multi-iteration session. These diagnostics make the agent's search process
auditable and show where the search stalls.

\begin{table}[t]
\centering
\small
\caption{\small{
Rubric for annotating each agent iteration. Labels are assigned with LLMs (Claude Opus 4.7).
}}
\label{tab:rubric}
\vspace{-0.5em}
\renewcommand{\arraystretch}{1.15}
\resizebox{0.95\linewidth}{!}{%
\begin{tabular}{p{0.22\linewidth}p{0.36\linewidth}p{0.34\linewidth}}
\toprule
\rowcolor{rubricHeader}
\textbf{Label} & \textbf{Criterion} & \textbf{Example} \\
\midrule
\rowcolor{rubricRowA}
New policy family
& Moves to a different type of data strategy.
& Moves from length filtering to training-dynamics scoring. \\

\rowcolor{rubricRowB}
Grounded
& Justified by concrete evidence.
& Uses evaluation errors, data audits, prior runs, literature, or ablations. \\

\rowcolor{rubricRowA}
Effective
& Improves the target metric under the same budget and training recipe.
& Beats the previous policy under controlled comparison. \\

\rowcolor{rubricRowB}
\emph{Shallow}
& Makes a reactive local adjustment with weak rationale.
& ``This helped, so add more,'' without a task-specific hypothesis. \\
\bottomrule
\end{tabular}}
\renewcommand{\arraystretch}{1.0}
\vspace{-1.5em}
\end{table}

\section{Default Agents: Useful Executors, Narrow Researchers}\label{sec:prototype}
\subsection{Open-Prompt Agent Setup}


We evaluate generalist coding agents in the \bench{} environment under the
minimal scaffold with open-prompt. The agent receives
only the benchmark interface: the task objective, workspace, constraints, and
commands for validation, training, and evaluation. It must decide on its own
what data policy to implement and how to revise that policy across iterations.
It follows the standard inspect--curate--train--evaluate--log loop (Prompt text in Appendix~\ref{app:prompts}):

\begin{mdframed}[backgroundcolor=gray!5, linewidth=0.5pt]
\small{\textbf{Open-prompt loop.}
\begin{enumerate}[leftmargin=1.2em, itemsep=2pt]
\item Read the task brief and constraints.
\item Inspect the candidate pool and previous runs.
\item Propose a data policy, such as selection, mixing, filtering, scoring, or rewriting.
\item Materialize a training set and save a manifest.
\item Fine-tune the model with the fixed training recipe.
\item Evaluate on the fixed benchmark suite.
\item Write a run note explaining the hypothesis, evidence, and next action.
\end{enumerate}}
\end{mdframed}
This setup gives the agent procedural access to the curation loop while leaving
the choice of data policy to the agent.

\begin{table}[h]
\centering
\small\vspace{0.2em}
\caption{\small{Generalist agents (\textbf{open-prompt}) selecting 10k samples from LLaVA-665K dataset for training LLaVA-1.5-7B models. Scores are averaged over 8
benchmarks; ``\% of full gain'' is the fraction of the
base$\to$full-data fine-tuning improvement recovered. \textcolor{Green}{Green} values are
deltas over the best random run. Best outcomes are marked in \textbf{bold}. All agents perform better than the best of random while meeting or exceeding evaluated human baselines.}}\vspace{-0.5em}
\label{tab:main}
\resizebox{\linewidth}{!}{
\begin{tabular}{l|lll}
\toprule
\textbf{Setting} & \textbf{Avg. score} \tiny{$\pm$std} & \textbf{[min, max]} & \textbf{\% of full gain} \\
\midrule
Base model (no fine-tuning) & 28.8 & - & 0\% \\
Full-data fine-tuning (665k) & 37.1 & - & 100\% \\
\midrule
\multicolumn{4}{c}{\textit{Baselines, 10k budget}} \\
\midrule
Random (baseline, 10 runs) & 31.9\tiny{$\pm$0.3} & [31.5, \textbf{32.5}] & 45\% (best of 10)\\
ICONS \citep{wu2024icons}  & 33.3 & - & 54\% \\
ARDS \citep{yang2025data}  & 33.2 & - & 52\% \\
\midrule
\multicolumn{4}{c}{\textit{Open-prompt Agents, 10k budget, 10 iterations}} \\
\midrule
Claude Code (Opus 4.7) & \textbf{33.7}\tiny{$\pm$0.3} \small{\textbf{\textcolor{Green}{(+1.2)}}} & [32.5, \textbf{34.0}] & \textbf{59\% \textcolor{Green}{(+14\%)}} \\
Codex (GPT-5.4) & 33.3\tiny{$\pm$0.2} \small{\textcolor{Green}{(+0.8)}} & [33.0, 33.5] & 54\% \textcolor{Green}{(+9\%)} \\
Kimi K2.5 (OpenHands) & 32.8\tiny{$\pm$0.5} \small{\textcolor{Green}{(+0.3)}} & [32.2, 33.4] & 48\% \textcolor{Green}{(+3\%)} \\
Qwen3.5-397B (OpenHands) & 33.2\tiny{$\pm$0.2} \small{\textcolor{Green}{(+0.7)}} & [33.0, 33.4] & 53\% \textcolor{Green}{(+8\%)} \\
\bottomrule
\end{tabular}}\vspace{-1em}
\end{table}

\begin{table*}[t]
\centering
\small
\caption{\small{
Example open-prompt trajectory on the 10k LLaVA-665K selection task using
Claude Code. Scores are averaged over 8 VLM benchmarks after LLaVA-1.5-7B
fine-tuning. After an initial subset-balancing policy, most later edits adjust
source ratios rather than switch policy families, illustrating the
local-exploitation pattern in Finding~2.
}}
\label{tab:unscaf-10-res}
\vspace{-0.5em}
\resizebox{0.93\linewidth}{!}{%
\renewcommand{\arraystretch}{1.12}
\begin{tabular}{C{0.45cm}C{1.15cm}C{0.95cm}p{6.2cm}C{1.15cm}C{1.15cm}C{1.15cm}C{1.05cm}}
\toprule
\textbf{\#} & \textbf{Score} & \textbf{Status} & \textbf{Policy description} 
& \textbf{New} & \textbf{Grounded} & \textbf{Effective} & \textit{\textbf{Shallow}} \\
\midrule
\rowcolor{baselineRow}
-- & 28.95 & keep & Pre-init baseline, no fine-tuning & -- & -- & -- & -- \\
\rowcolor{baselineRow}
-- & 32.38 & keep & Random 10k baseline & -- & -- & -- & -- \\
\midrule
1 & 33.06 & keep 
& Balanced subset; OCR boost; no text\_only 
& \yes & \yes & \yes & \no \\

\rowcolor{discardRow}
2 & 32.88 & discard 
& Balanced subsets + longest assistant responses 
& \yes & \no & \no & \yes \\

\rowcolor{discardRow}
3 & 32.06 & discard 
& Balanced subsets + text\_only (500) 
& \no & \no & \no & \yes \\

4 & 33.23 & keep 
& Equal allocation: 2k per visual subset 
& \no & \no & \yes & \yes \\

\rowcolor{discardRow}
5 & 32.56 & discard 
& Boost VG/GQA; reduce COCO/OCR 
& \no & \no & \no & \yes \\

6 & 33.63 & keep 
& Near-equal mix: COCO=2500, others=1875 
& \no & \yes & \yes & \no \\

\rowcolor{discardRow}
7 & 32.95 & discard 
& COCO=3000, others=1750 
& \no & \yes & \no & \no \\

8 & 33.70 & keep 
& OCR-boosted mix: COCO=2000, OCR/Text=2250 
& \no & \yes & \yes & \no \\

\rowcolor{discardRow}
9 & 33.02 & discard 
& Heavy OCR+VG: COCO=1500 
& \no & \yes & \no & \yes \\

\rowcolor{bestRow}
10 & \textbf{33.74} & keep 
& GQA+TextVQA boost: GQA=2250, TextVQA=2250 
& \no & \yes & \yes & \no \\
\midrule
\multicolumn{4}{@{}l}{\textbf{Execution success:} 10/10 \hspace{2em} \textbf{Trajectory totals}} 
& 2/10 & 6/10 & 5/10 & 5/10 \\
\bottomrule
\end{tabular}
\renewcommand{\arraystretch}{1.0}
}
\vspace{-1em}
\end{table*}
\subsection{Outcome: Generalist Agents Match Human-Designed Baselines}
We score each agent session using the outcome and trajectory metrics defined in \Cref{sec:bench_diagnostics}. Experimental details, additional results, and ablation studies are provided in Appendix~\ref{app:exps}.

We also report the rate of \textbf{Execution success:} whether the agent completes the required loop without violating task constraints---producing a valid curated dataset, interacting with the benchmark, and logging the result. We find that execution is reliable in this setting: across more than 500
iterations in 50+ sessions, agents produced fewer than 10 crashed iterations not
attributable to external incidents, and these crashes did not derail the
corresponding sessions.


\vspace{-0.5em}\paragraph{Finding 1: Generalist agents substantially improve training outcomes through data curation, reaching expert-baseline performance in a few iterations.} 
Figure~\ref{fig:main_4in1} shows that the benefits of open-prompt agentic curation extend beyond the primary LLaVA fine-tuning setting. Across multiple fine-tuning tasks (LLaVA-1.5-7B on LLaVA-665K, SmolVLM-Base on LLaVA-665K, and SmolVLM-Base on Vision-Flan), agents consistently improve over random selection. The same pattern also appears in pretraining: on the DataComp Small CLIP-pretraining task, the agent achieves a clear gain over the strongest filtering baseline, top-30\% CLIP L/14-score filtering \citep{brunner2023datacomp}, showing that the agentic curation loop is not limited to instruction-tuning data. \Cref{tab:main} compares generalist agents with random selection and
published data-selection baselines on the primary LLaVA-665K/LLaVA-1.5-7B task. Here, all open-prompt agents improve over the best random run on average. Claude Code performs best, reaching $33.7$ versus $32.5$ for the best random run and $33.3$/$33.2$ for ICONS~\cite{wu2024icons}/ARDS~\cite{yang2025data}; this recovers $59\%$ of the full-data gain using only $\approx 1.5\%$ of the 665k pool. Codex and Qwen are close to the 10k human-designed baselines, while Kimi improves over random but remains below them.

\subsection{Trajectory: Success Comes from Local Heuristics}
\paragraph{Finding 2: Open-prompt agents make weakly grounded choices and over-exploit local policy families.} Despite favorable final scores, open-prompt agents show two recurring failure
modes. First, their stated rationales are often weakly grounded. Run notes
typically invoke plausible but generic goals---``filter low-quality examples,''
``increase diversity,'' ``try different random seeds,'' or ``balance
domains''---without tying the next policy to a concrete evidence source,
hypothesis, or ablation (Tables~\ref{tab:unscaf-10-res},\ref{tab:codex-base10}).
This produces a form of \emph{vibe optimization}: the plan sounds like data
research, but the operational criterion is underspecified.

Second, when the agent commits to a manifest, it usually instantiates the
lowest-cost version of the idea: a source-ratio change, response-length
threshold, metadata filter, or seed change. These local moves might still improve
the score, but they keep the search within the same policy family. Once the
agent finds a workable family, it rarely pivots to another, and the final outcome is bounded by that family 
(Tables~\ref{tab:unscaf-10-res},\ref{tab:scaffold-summary}).




\section{Scaffolds Convert Execution into Exploration}
\label{sec:scaffolds}

The previous section shows that open-prompt agents are useful executors but
narrow researchers: they can run the loop and improve scores, yet they tend to
settle into local data-policy edits. We now introduce a scaffold
ladder with different levels of guidance to change this behavior.
The key question is whether guidance merely changes what agents write in their
plans, or whether it changes the policies they actually execute.

\newcolumntype{L}[1]{>{\raggedright\arraybackslash}p{#1}}

\begin{table*}[t!]
\centering
\begin{minipage}{0.73\textwidth}
\centering
\footnotesize
\caption{\small{
Scaffold conditions. Light scaffolds provide optional guidance; heavy scaffolds
make research discipline part of the task contract.
}}
\label{tab:prompt-excerpts}
\vspace{-0.5em}
\setlength{\tabcolsep}{4pt}
\renewcommand{\arraystretch}{1.18}
\resizebox{\linewidth}{!}{%
\begin{tabular}{L{2.35cm}L{1.65cm}L{2.75cm}L{5.45cm}}
\toprule
\rowcolor{rubricHeader}
\textbf{Condition} &
\textbf{Constraint} &
\textbf{What the agent must do} &
\textbf{Operative prompt language} \\
\midrule

\rowcolor{rubricgray}
\textbf{Open-prompt} &
None &
Run the standard curation loop &
Select at most $B$ examples; inspect data, write scripts, submit a manifest,
read logs, and iterate. \\

\midrule
\rowcolor{rubricblue}
\textbf{Light I}\newline Data Strategies &
Optional &
Consider broader policy families &
Curation strategies include quality filtering, source balancing, semantic
diversity, task balancing, deduplication, and length/format filters. \\

\rowcolor{rubricblue}
\textbf{Light II}\newline Research Papers &
Optional &
Consult prior-method skill cards &
List available paper-derived skill cards and \textbf{consider using} them when
helpful. \\

\midrule
\rowcolor{bestRow}
\textbf{Heavy I}\newline Self-Research &
Binding &
Justify each run with evidence &
Before editing the policy, write an \textbf{observation}, \textbf{hypothesis},
expected effect, and minimal change. The observation \textbf{must cite concrete
evidence}. \\

\rowcolor{bestRow}
\textbf{Heavy II}\newline Adapt Papers &
Binding &
Adapt a prior method into code &
For every non-baseline iteration, \textbf{choose a prior method}, explain why it
applies, \textbf{adapt} it to the available fields and budget, implement the
policy, and \textbf{validate} the manifest. \\

\bottomrule
\end{tabular}}
\renewcommand{\arraystretch}{1.0}
\vspace{-1em}
\end{minipage}
\end{table*}

\vspace{-0.5em}\subsection{Scaffold Design}

We study two light scaffolds and two heavy scaffolds. The light scaffolds are
designed as awareness-level interventions. \emph{Data strategies} gives the
agent a short list of possible policy families, including source balancing,
quality filtering, diversity sampling, task targeting, deduplication,
difficulty scoring, conversation-structure filtering, and image-metadata
filtering. \emph{Research papers} gives the agent access to paper-derived
skill cards and asks it to consult them when useful. These scaffolds expand the
agent's stated option set, but do not require a specific decision protocol.

The heavy scaffolds instead impose a binding research protocol. In
\emph{self-research}, each non-baseline iteration must begin with a written
observation, hypothesis, expected metric effect, and minimal policy change,
grounded in prior runs or evaluation logs. In \emph{adapt papers}, each
non-baseline iteration must cite a specific prior method or skill card,
explain why it applies, adapt it to the available fields and budget, and
validate the resulting manifest before training. The scaffold constrains the
structure of the research loop, but not the policy content: the agent still
chooses which failure mode to address, which method to adapt, and how to
implement the adaptation.

Essentially, light scaffolds test whether
making the solution space visible is enough; heavy scaffolds test whether the
agent needs a procedural constraint that forces evidence-grounded or
method-grounded execution.

\vspace{-0.5em}
\subsection{Intervention Outcome} 
\begin{figure}[t]
\centering
\includegraphics[width=\linewidth]{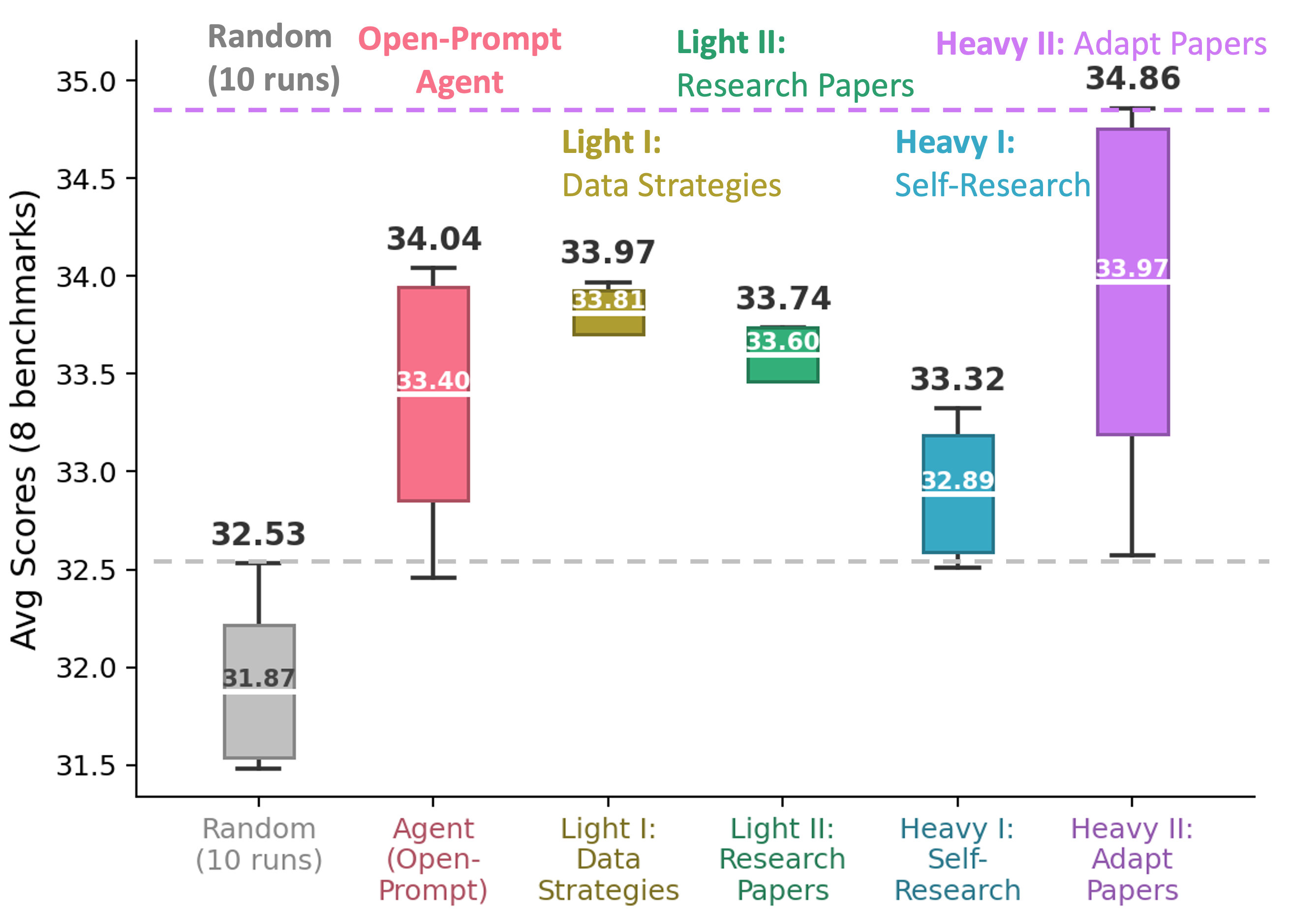}
\vspace{-2em}
\caption{\small{
Agentic data-curation results with different scaffolds (Claude Code, 10-iteration). Fine-tuning pre-trained LLaVA-1.5-7B on curated subsets of 10k samples from the LLaVA-665K instruction-tuning pool and evaluating on 8 benchmarks. Scaffolding significantly alters agents' behavior and outcomes. Compared to \textbf{open-prompt}, \textbf{light} scaffolds reduce outcome variances but do not improve the best outcome. \textbf{Heavy} scaffolds change the outcome substantially, but the effects may be for the better or for the worse depending on the specific scaffold design. 
}}
\vspace{-1.5em}
\label{fig:main_scaffolds}
\end{figure}

Figure~\ref{fig:main_scaffolds} and Table~\ref{tab:scaffold-summary} summarize
the scaffold results. The main pattern is that awareness-level scaffolds
broaden what the agent considers, but protocol-level scaffolds change what the
agent executes. Light scaffolds reduce variance and increase some trajectory
quality metrics, but they do not materially improve the best outcome over the
open-prompt condition. The strongest result comes from the paper-adaptation
scaffold: its best 10k policy reaches 34.9 average score, exceeding the
open-prompt agent and the evaluated 100k ARDS baseline while using one-tenth
as many examples.

\newcolumntype{P}[1]{>{\raggedright\arraybackslash}m{#1}}
\newcolumntype{Q}[1]{>{\centering\arraybackslash}m{#1}}

\definecolor{scaffoldHeader}{RGB}{232,240,248}
\definecolor{scaffoldGroup}{RGB}{248,248,248}
\definecolor{scaffoldLight}{RGB}{243,248,253}
\definecolor{scaffoldHeavy}{RGB}{244,250,244}
\definecolor{scaffoldBest}{RGB}{226,244,229}

\begin{table*}[t!]
\centering
\footnotesize
\caption{\small{
Effect of scaffold design on Claude Code for 10k-example selection from
LLaVA-665K to fine-tune LLaVA-1.5-7B. Outcome scores are averaged over eight
benchmarks; trajectory diagnostics report the fraction of iterations with each
annotation. Green deltas are relative to open-prompt. 
}}
\label{tab:scaffold-summary}
\vspace{-0.5em}
\setlength{\tabcolsep}{3.5pt}
\renewcommand{\arraystretch}{1.23}
\resizebox{0.95\linewidth}{!}{%
\begin{tabular}{P{2.45cm}Q{1.25cm}Q{1.55cm}P{3.75cm}Q{1.5cm}Q{1.6cm}Q{1.5cm}Q{1.5cm}}
\toprule
\rowcolor{scaffoldHeader}
\textbf{Condition} &
\multicolumn{3}{c}{\textbf{Outcome}} &
\multicolumn{4}{c}{\textbf{Trajectory diagnostics}} \\
\cmidrule(lr){2-4}\cmidrule(l){5-8}
\rowcolor{scaffoldHeader}
&
\textbf{Mean} &
\textbf{Range} &
\textbf{Best executed policy} &
\textbf{New $\uparrow$} &
\textbf{Grounded $\uparrow$} &
\textbf{Effective $\uparrow$} &
\textbf{Shallow $\downarrow$} \\
\midrule

\rowcolor{rubricgray}
\textit{Open-prompt}\newline
\textit{baseline} &
\makecell[c]{\textit{33.7}\\\textit{59\% gain}} &
\textit{[32.5, 34.0]} &
Source-ratio tuning: boost GQA and TextVQA &
\makecell[c]{\textit{27\%}\\[-1pt]{\scriptsize$\pm$12}} &
\makecell[c]{\textit{57\%}\\[-1pt]{\scriptsize$\pm$6}} &
\makecell[c]{\textit{\textbf{30\%}}\\[-1pt]{\scriptsize$\pm$20}} &
\makecell[c]{\textit{47\%}\\[-1pt]{\scriptsize$\pm$6}} \\

\midrule
\multicolumn{8}{@{}l}{\textbf{Light scaffolds: optional guidance}} \\

\rowcolor{scaffoldLight}
\textbf{Data Strategies}\newline
Light I &
\makecell[c]{\textbf{33.8}\\\textbf{60\% gain}} &
[\textbf{33.7}, 34.0] &
8k COCO-heavy source mixture &
\makecell[c]{\textbf{43\%}\\[-1pt]{\scriptsize$\pm$12\%}\\[-1pt]\textcolor{Green}{\scriptsize(+16\%)}} &
\makecell[c]{51\%\\[-1pt]{\scriptsize$\pm$10\%}} &
\makecell[c]{23\%\\[-1pt]{\scriptsize$\pm$12\%}} &
\makecell[c]{46\%\\[-1pt]{\scriptsize$\pm$5\%}} \\

\rowcolor{scaffoldLight}
Research Papers\newline
Light II &
33.6 &
[33.4, 33.7] &
Scaled-up source proportions from the best prior run &
\makecell[c]{38\%\\[-1pt]{\scriptsize$\pm$6\%}} &
\makecell[c]{\textbf{70\%}\\[-1pt]{\scriptsize$\pm$11\%}\\[-1pt]\textcolor{Green}{\scriptsize(+13\%)}} &
\makecell[c]{27\%\\[-1pt]{\scriptsize$\pm$9\%}} &
\makecell[c]{\textbf{37\%}\\[-1pt]{\scriptsize$\pm$19\%}\\[-1pt]\textcolor{Green}{\scriptsize(-10\%)}} \\

\midrule
\multicolumn{8}{@{}l}{\textbf{Heavy scaffolds: binding research protocol}} \\

\rowcolor{scaffoldHeavy}
Self-Research\newline
Heavy I &
32.9 &
[32.5, 33.3] &
Remove text\_only/VG; shift small OCR\_VQA quota to COCO &
\makecell[c]{34\%\\[-1pt]{\scriptsize$\pm$5\%}} &
\makecell[c]{91\%\\[-1pt]{\scriptsize$\pm$10\%}} &
\makecell[c]{23\%\\[-1pt]{\scriptsize$\pm$8\%}} &
\makecell[c]{9\%\\[-1pt]{\scriptsize$\pm$10\%}} \\

\rowcolor{scaffoldBest}
\textbf{Adapt Papers}\newline
\textbf{Heavy II} &
\makecell[c]{\textbf{34.0}\\\textbf{63\% gain}} &
\makecell[c]{[32.6, \textbf{34.9}]\\[-1pt]\textcolor{Green}{\scriptsize \textbf{best:+0.9/+11\%}}} &
\textbf{EL2N top-loss selection + assistant-loss noise filtering} &
\makecell[c]{\textbf{67\%}\\[-1pt]{\scriptsize$\pm$6\%}\\[-1pt]\textbf{\textcolor{Green}{\scriptsize(+40\%)}}} &
\makecell[c]{\textbf{100\%}\\[-1pt]{\scriptsize$\pm$0\%}\\[-1pt]\textbf{\textcolor{Green}{\scriptsize(+43\%)}}} &
\makecell[c]{20\%\\[-1pt]{\scriptsize$\pm$0\%}} &
\makecell[c]{\textbf{0\%}\\[-1pt]{\scriptsize$\pm$0\%}\\[-1pt]\textbf{\textcolor{Green}{\scriptsize(-47\%)}}} \\

\midrule
\multicolumn{8}{@{}p{0.98\textwidth}@{}}{
\textbf{Non-agent reference points.}
Random 100k: 33.7 / 59\% gain; Random 200k: 34.0 / 63\% gain;
ICONS 100k: 34.5 / 69\% gain~\citep{wu2024icons};
ARDS 100k: 34.1 / 64\% gain~\citep{yang2025data}.
} \\

\bottomrule
\end{tabular}}
\renewcommand{\arraystretch}{1.0}
\vspace{-1.2em}
\end{table*}

\vspace{-0.5em}
\paragraph{Finding 3: Light scaffolds broaden consideration, but do not reliably improve executed policy quality.}
The data-strategies scaffold increases the rate of new-policy-family moves
from $27\%$ to $43\%$, suggesting that the agent does consider a broader
space when policy families are surfaced explicitly. The research-paper
scaffold increases grounding from $57\%$ to $70\%$ and reduces shallow moves
from $47\%$ to $37\%$, indicating non-trivial trajectory changes. However, neither
light scaffold improves the best outcome beyond the open-prompt maximum of
$34.0$. The data-strategies condition reaches the same maximum, and the
research-paper condition remains below it.

The reason is that light scaffolds often change the agent's vocabulary more
than its implementation. The agent names richer policy families, but still
tends to instantiate low-cost variants: quota changes, length filters,
simple quality proxies, or scaled versions of previously successful mixtures (\Cref{tab:cc-light-1},\ref{tab:cc-light-2}).
This is a rational short-horizon behavior. Once the agent finds a workable
family from prior
runs, local edits preserve information from prior
runs and have a reasonable chance of producing small gains. Light scaffolds make
different policy options visible, but do not force the agent to pay the implementation
cost.


\vspace{-0.5em}
\paragraph{Finding 4: heavy scaffolds enforce grounding and enable larger breakthroughs.} The heavy scaffolds change the executed trajectory more substantially.
Self-research increases grounded iterations from $57\%$ to $91\%$ and reduces
shallow operations from $47\%$ to $9\%$. Adapt-papers is stronger: all
iterations are grounded, shallow operations are eliminated, and the rate of
new-policy-family moves rises to $67\%$. This indicates that the scaffold is
not merely changing the wording of the agent's notes; it changes what the agent
actually tries.

The effect is clearest in the best adapt-papers trajectory. After several
failed method adaptations, the agent moves into a training-dynamics family:
it uses an EL2N-style loss proxy to select high-loss examples, then refines
that policy with an assistant-loss noise filter to remove the highest-loss
samples likely to reflect incoherent or noisy responses. This produces the
best 10k result, $34.9$, above the open-prompt agent and the evaluated 100k baselines.

This result also explains why trajectory diagnostics are needed. Heavy
scaffolds do not necessarily maximize the number of locally effective
iterations. The adapt-papers scaffold has only $20\%$ effective iterations by
the immediate-improvement criterion, lower than the open-prompt condition.
But the trajectory reaches a policy family that open-prompt search did not
reach. For autonomous data research, the important quantity is therefore not
only the rate of small local gains, but whether the search reaches
higher-upside regions of the policy space.

The scaffold comparison should not be read as showing that heavier is always
better. Self-research improves grounding but lowers the final score vs. open-prompt. Heavy scaffolds combine several ingredients---method grounding,
structured logging, category rotation, and procedural constraints---so
the experiment does not isolate a single causal component. The more conservative
conclusion is that awareness alone is insufficient: to convert agent execution
into policy exploration, the scaffold must make evidence-grounded or
method-grounded action part of the task contract.

\vspace{-0.5em}\section{Broader Investigations}\vspace{-0.5em}
\subsection{Scaling the Curation Compute Budget}
\begin{figure}[h]
\centering
\includegraphics[width=\linewidth]{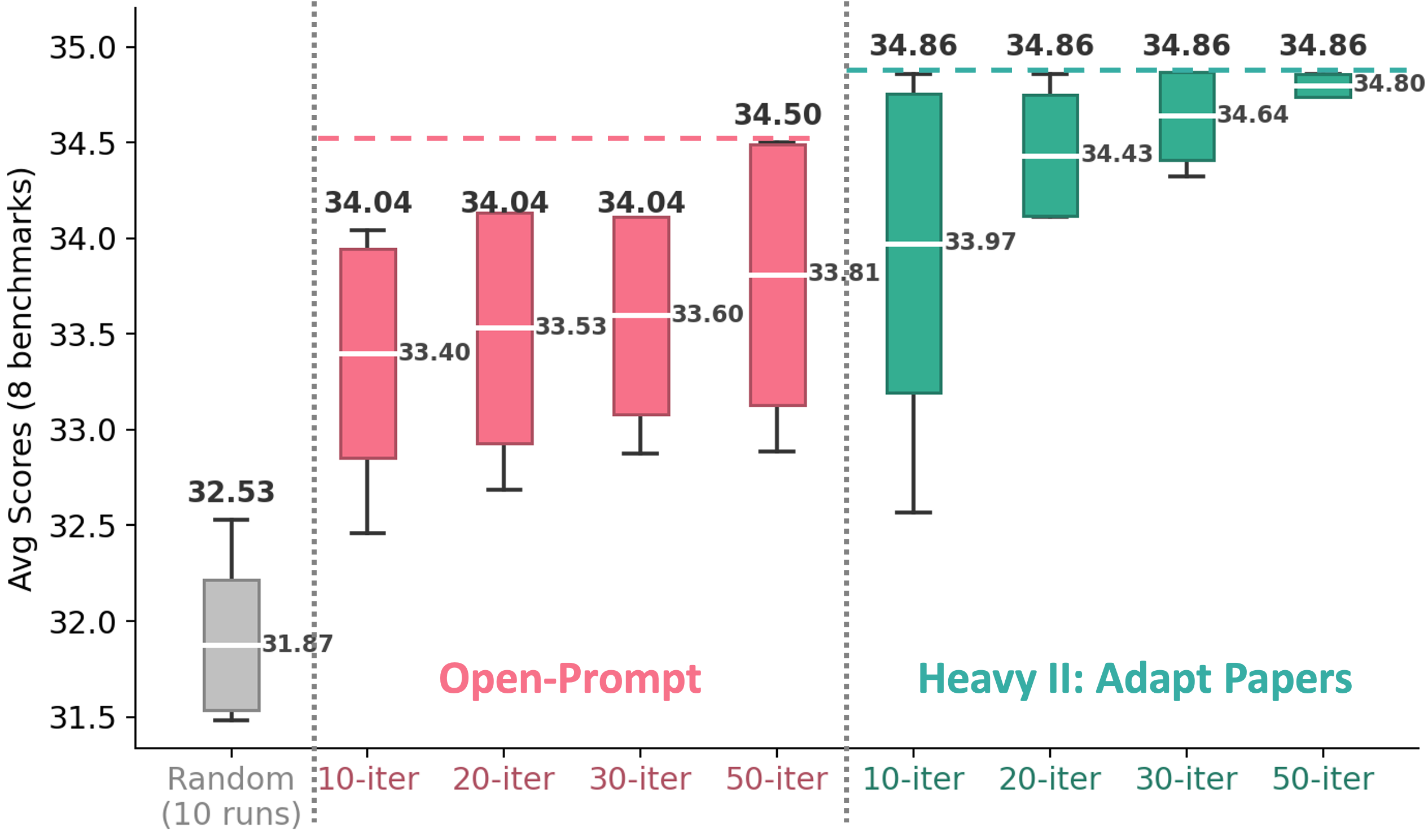}
\vspace{-1.5em}
\caption{\small{
Ablation on the number of iterations (10-50) in each agent session (Claude Code).
}}
\vspace{-1.3em}
\label{fig:main_res1}
\end{figure}
\paragraph{Finding 5: Longer curation search continues to pay off under a fixed data budget.} Figure~\ref{fig:main_res1} varies the number of agent iterations from 10 to
50 while keeping the candidate pool, data budget, training recipe, and
evaluation suite fixed. The result is notable: average performance continues to
improve within this range rather than clearly plateauing. Under open-prompting,
the gains accumulate gradually, consistent with extended local search over
source mixtures and related heuristics. Under the Heavy II scaffold, the best
score is already found within the first 10 iterations, but additional iterations
reduce variance and move the average upward.

This suggests that the agent iteration budget is itself a meaningful compute
axis for data curation. In a finite-data regime, where additional raw data is
unavailable, expensive, or lower quality, progress need not come only from
training on more examples. Additional compute can instead be spent on searching
over data policies: selecting different subsets, testing distinct policy
families, adapting prior methods, or rewriting examples. From this perspective,
\bench{} provides a way to study how curation quality scales with search
compute under a fixed data pool.

\vspace{-0.5em}\subsection{Rewriting Extension}\label{sec:rewrite}
The selection experiments optimize \emph{which} examples to train on. To test
whether \bench{} also supports richer data actions, we instantiate it as a
data-rewriting task: the agent selects examples, rewrites them with an external
multimodal large language model (MLLM) tool, verifies the rewritten samples, and submits the transformed dataset
through the same validation, contamination, training, and evaluation loop.
This setting uses a stricter \emph{self-research}-like scaffold. Every post-baseline
iteration must inspect raw evaluation failures, name a failure mode, select
source examples tied to that mode, rewrite them with the MLLM, verify the
rewritten examples, and pass contamination checks before training.  Full rewrite scaffold is provided in code repo.
 Results (\Cref{tab:rewrites}) show that \bench{} is not limited to subset selection: the same framework can evaluate richer data actions, such as targeted rewriting, when equipped with appropriate tools.

\begin{table}[h]
\centering
\small\vspace{-0.3em}
\caption{\small{Rewriting results on LLaVA-665K/LLaVA-1.5-7B with self-research-like rewriting scaffolding (Claude Code agent). The agent selects and rewrites 10k examples using an external MLLM tool. Green deltas show improvement over the \emph{Template Rewriting} baseline. Bottom rows provide data-efficiency reference points. The choice of rewriting model substantially affects outcomes; Qwen3.5-9B consistently outperforms smaller alternatives. }}\vspace{-0.5em}
\label{tab:rewrites}
\resizebox{\linewidth}{!}{
\begin{tabular}{l|l|ll}
\toprule
\textbf{Setting} & \textbf{Data size} & \textbf{Avg. score} & \textbf{\% of full gain}  \\
\midrule
Base model & 0 & 28.8  & 0\%  \\
Full-data fine-tuning & 665k & 37.1 & 100\%  \\
\midrule
Random (10-run max) & 10k & 32.5 & 45\% \\
\makecell[l]{\textit{Template Rewriting (baseline)}\\
- \textit{\citet{wang2024template}}}  & \textit{10k} & \textit{32.6} & \textit{46\%}\\
\midrule
\makecell[l]{Agent (tool: Qwen3.5-2B): 10-iter}  & 10k & 33.4 \textcolor{Green}{(+0.8)}  & 55\% \textcolor{Green}{(+9\%)}\\
\makecell[l]{\textbf{Agent (tool: Qwen3.5-9B)}: 10-iter}  & 10k & \textbf{34.1  \textcolor{Green}{(+1.5)}} & \textbf{64\% \textcolor{Green}{(+18\%)}} \\
\makecell[l]{Agent (tool: Gemma3-4B): 10-iter} & 10k & 32.8 \textcolor{Green}{(+0.2)} & 48\% \textcolor{Green}{(+2\%)}\\
\midrule
\makecell[l]{Agent (tool: Qwen3.5-2B): 15-iter}  & 10k & 34.0 \textcolor{Green}{(+1.4)} & 63\% \textcolor{Green}{(+17\%)}\\
\makecell[l]{\textbf{Agent (tool: Qwen3.5-9B)}: 20-iter}  & 10k & \textbf{34.7 \textcolor{Green}{(+2.1)}}& \textbf{71\% \textcolor{Green}{(+25\%)}}\\
\makecell[l]{Agent (tool: Gemma3-4B): 15-iter}  & 10k & 33.1 \textcolor{Green}{(+0.5)} & 52\% \textcolor{Green}{(+6\%)}\\
\makecell[l]{Agent (tool: Gemma3-4B): 25-iter} & 10k & 33.6 \textcolor{Green}{(+1.0)}  & 58\% \textcolor{Green}{(+12\%)}\\
\midrule
Random 50k & 50k & 33.4\tiny{$\pm$0.2}  & 55\%\tiny{$\pm$3\%}  \\
Random 100k & 100k & 33.7\tiny{$\pm$0.2}& 59\%\tiny{$\pm$3\%} \\
Template Rewriting 50k & 50k & 33.5 & 57\% \\
Template Rewriting 100k & 100k & 33.8 & 60\%\\
\bottomrule
\end{tabular}}\vspace{-1.1em}
\end{table}

\subsection{Comparisons with Broader Data Research}
In addition, we compare prominent data selection from existing research for a broader comparison with the agents.  Different from experiments in Section ~\ref{sec:prototype}, which compares with human-designed methods originally proposed for these tasks, we now expand the comparision and include broader methods that were originally proposed for generic use cases or different applications. All these methods are implemented from scratch, adapted for new tasks and targeting the same validation goals as the data agents.

We compare with representative approaches, LESS~\citep{xia2024less}, DSIR~\citep{xie2023data}, and D4~\citep{tirumala2023d4}, to achieve a good coverage of distinct mythologies. Specifically, LESS~\citep{xia2024less} selects instruction tuning data via estimated influence on the validation tasks. It requires first running an expensive training session to evaluate the influence of every candidate example on the target tasks, which scales with the size of the total candidate data. DSIR~\citep{xie2023data} selects data for pre-training/continued pre-training by matching the distribution to the target tasks. It operates on the n-gram distribution and selects data directly targeting validation tasks. D4~\citep{tirumala2023d4} conducts density-based deduplication for pre-training-scale data filtering/selection. This method works independently of the validation tasks. 
We conduct 3 independent runs for each of these methods (selection, training, evaluation) and report their results 
on selecting from LLaVA-665K to fine-tune LLaVA-1.5-7B
and DataComp Small benchmark (Filtering
track) in Table~\ref{tab:additional_results_1} and ~\ref{tab:additional_results_2}, respectively.

\begin{table}[h]
\centering
\small\vspace{-0.3em}
\caption{\small{Comparisons with broader data selection methods (LESS, DSIR, D4) on selecting 10k samples from LLaVA-665K to fine-tune LLaVA-1.5-7B, evaluated on 8 benchmarks. ``\% of full gain'' is the fraction of the base$\rightarrow$full-data fine-tuning improvement recovered. \textcolor{Green}{Green} values denote deltas over the random baseline (37\%) by more than 10\% in "\% of full gain". Best outcomes in each catgory are marked in \textbf{bold}.}}
\vspace{-0.5em}
\label{tab:additional_results_1}
\resizebox{\linewidth}{!}{
\begin{tabular}{l|ll}
\toprule
\textbf{Setting} & \textbf{Avg. score} \tiny{$\pm$std} & \textbf{\% of full gain} \\
\midrule
Base model (no fine-tuning) & 28.8 & 0\% \\
Full-data fine-tuning (665k) & 37.1 & 100\% \\
\midrule
\multicolumn{3}{c}{\textit{Baselines, 10k budget}} \\
\midrule
Random (baseline, 10 runs) & 31.9\tiny{$\pm$0.3} & 37\% \\
ICONS & 33.3 & 54\% \textcolor{Green}{(+17\%)} \\
ARDS & 33.2 & 53\% \textcolor{Green}{(+16\%)} \\
\textbf{LESS} & \textbf{33.6\tiny{$\pm$0.3}} \textbf{\textcolor{Green}{(+1.7)}} & \textbf{58\% \textcolor{Green}{(+20\%)}} \\
DSIR & 32.5\tiny{$\pm$0.2} \small{(+0.6)} & 45\% \textcolor{Green}{(+8\%)} \\
D4 & 32.0\tiny{$\pm$0.3} \small{(+0.1)} & 39\% (+2\%) \\
\midrule
\multicolumn{3}{c}{\textit{Open-prompt Agents, 10k budget, 10 iterations}} \\
\midrule
\textbf{Claude Code (Opus 4.7)} & \textbf{33.7\tiny{$\pm$0.3} \small{\textcolor{Green}{(+1.8)}}} & \textbf{59\% \textcolor{Green}{(+22\%)}} \\
Codex (GPT-5.4) & 33.3\tiny{$\pm$0.2} \small{\textcolor{Green}{(+1.4)}} & 54\% \textcolor{Green}{(+17\%)} \\
Kimi K2.5 (OpenHands) & 32.8\tiny{$\pm$0.5} \small{\textcolor{Green}{(+0.9)}} & 48\% \textcolor{Green}{(+11\%)} \\
Qwen3.5-397B (OpenHands) & 33.2\tiny{$\pm$0.2}  \small{\textcolor{Green}{(+1.3)}} & 53\% \textcolor{Green}{(+16\%)} \\
\midrule
\multicolumn{3}{c}{\textit{Claude+Scaffolds, 10k budget, 10 iterations}} \\
\midrule
Light I, Data Strategies & 33.8\tiny{$\pm$0.1} \small{\textcolor{Green}{(+1.9)}} & 60\% \textcolor{Green}{(+23\%)} \\
Light II, Research Papers & 33.6\tiny{$\pm$0.1} \small{\textcolor{Green}{(+1.7)}} & 58\% \textcolor{Green}{(+20\%)} \\
Heavy I, Self-Research & 32.9\tiny{$\pm$0.3} \small{\textcolor{Green}{(+1.0)}} & 49\% \textcolor{Green}{(+12\%)} \\
\textbf{Heavy II, Adapt Papers} & \textbf{34.0\tiny{$\pm$0.8} \small{\textcolor{Green}{(+2.1)}}} & \textbf{63\% \textcolor{Green}{(+26\%)}} \\
\bottomrule
\end{tabular}}
\vspace{-1.1em}
\end{table}

\begin{table}[h]
\centering
\small
\caption{\small{Comparisons with broader data selection methods
(DSIR, D4) on the DataComp Small benchmark (Filtering track). We pre-train ViT-B/32 CLIP models and report average scores across 38 tasks. We ran DSIR and D4 each for 3 different filtering budgets (retaining 70\%/50\%/30\% data). \% of "Full data" is the fraction of the achieved performance relative to the score from training on full data. \textcolor{Green}{Green} values denote improvements over the Random 50\% baseline. Best outcomes in each category are marked in \textbf{bold}.}}
\vspace{-0.5em}
\label{tab:additional_results_2}
\resizebox{\linewidth}{!}{
\begin{tabular}{l|ll}
\toprule
\textbf{Setting} & \textbf{Avg. score} \tiny{$\pm$std} & \textbf{\% of "Full data"} \\
\midrule
Full data & 0.134\tiny{$\pm$0.001} & 100\% \\
\midrule
\multicolumn{3}{c}{\textit{Baselines, 10 iterations}} \\
\midrule
Random 50\% data & 0.125\tiny{$\pm$0.001} & 93\% \\
\makecell[l]{\textbf{CLIP score filtering}\\
\textbf{(L/14-top 30\% data,}\\
\textbf{best DataComp baseline)}}
& \textbf{0.147\tiny{$\pm$0.003}
\small{\textcolor{Green}{(+0.022)}}}
& \textbf{110\% \textcolor{Green}{(+17\%)}} \\
\midrule
DSIR (retaining 70\% data) & 0.125\tiny{$\pm$0.002} & 93\% \\
DSIR (retaining 50\% data) & 0.123\tiny{$\pm$0.003} & 92\% \\
DSIR (retaining 30\% data) & 0.108\tiny{$\pm$0.005} & 81\% \\
\textbf{D4 (retaining 70\% data)} & \textbf{0.131\tiny{$\pm$0.002}
\small{\textcolor{Green}{(+0.006)}}} & \textbf{98\% \textcolor{Green}{(+5\%)}} \\
D4 (retaining 50\% data) & 0.125\tiny{$\pm$0.001} & 93\% \\
D4 (retaining 30\% data) & 0.109\tiny{$\pm$0.002} & 81\% \\
\midrule
\multicolumn{3}{c}{\textit{Open-prompt Agents, 10k budget, 10 iterations}} \\
\midrule
\textbf{Claude Code (Opus 4.7)}
& \textbf{0.166\tiny{$\pm$0.004}
\small{\textcolor{Green}{(+0.041)}}}
& \textbf{124\% \textcolor{Green}{(+31\%)}} \\
\bottomrule
\end{tabular}}
\vspace{-0.5em}
\end{table}

These methods are implemented directly in the same scenario and targeting the same validation tasks. LESS consumes 10x GPU hours for selection compared to the final training, spending the same amount of compute as 10 agent iterations. For the DataComp/CLIP experiment, DSIR and D4 each select for 3 different budgets to search for their best threshold. With all the confounding factors excluded, the open-prompt agent still reaches a performance comparable to the best human baselines, while the scaffold Heavy II, Adapt Papers could further surpass it.

Besides, the implementation of such human-designed methods is often highly specific — adapting them to a different task is nontrivial, and their effectiveness in extended scenarios is unknown a prior. This highlights a unique benefit and a major advantage of the data agents--these generalist agents provide an automated pipeline rather than a fixed solution. The same agent can be plugged into entirely different scenarios with minimal modifications. Beyond the data selection task for multimodal instruction tuning, the same agents work comparably well in distinct tasks, such as rewriting or selection for DataComp/CLIP training.

\section{Conclusion}\label{sec:conclusion}
This paper introduced \bench{}, a benchmark for evaluating whether
generalist agents can conduct training-data curation as an iterative
policy-search process. By fixing the model, training recipe, and
evaluation suite, \bench{} isolates the curated data policy as the
object of optimization and records the full trajectory of agent
decisions.

Our results show both promise and a limitation. Open-prompt agents are
already useful data-curation executors: they produce valid datasets,
run the training/evaluation loop, and match strong human-designed
selection baselines at small data budgets. However, trajectory
diagnostics reveal an execution--research gap: agents often improve
scores through local heuristic edits rather than by exploring distinct
policy families. Scaffold design is the central lever we identify for
narrowing this gap. Awareness-level scaffolds broaden plans but do not
reliably change executed policies; method-adaptation scaffolds force
agents to operationalize prior work and can open higher-upside regions
of the data-policy space.

\normalsize These findings suggest that future evaluations of autonomous data
research should report not only final scores, but also scaffold
strength and trajectory diagnostics: open-ended instructions and paper-grounded protocols answer different scientific questions. 

\clearpage
\section*{Limitations}
Our experiments instantiate \bench{} primarily in vision-language
instruction tuning, with one smaller CLIP-style pretraining instantiation
and a rewriting extension. The conclusions may differ for
pretraining mixtures, code, math, domain-specific instruction tuning, or
settings where evaluation is less reliable. Our scaffold comparison is also
not a full factorial study: heavy scaffolds combine method grounding, structured
logging, and stronger procedural constraints, so we identify a useful design
pattern rather than isolating a single causal ingredient. Finally, trajectory
labels require annotation; although they make agent behavior more legible, they
introduce their own judgment calls and should be reported with rubrics and
examples.

\section*{Broader Impacts}
Agentic data curation could reduce the cost of fine-tuning by identifying
smaller, higher-value subsets, but it also increases the need for oversight.
Agent-curated datasets may amplify biases, remove underrepresented examples,
or optimize toward narrow evaluation targets if deployed without review.
Because paper-grounded scaffolds lower the engineering barrier for data
optimization, they could also accelerate harmful applications. We therefore
recommend treating agent-curated datasets as candidate artifacts subject to
human or programmatic review, reporting scaffold strength and trajectory
diagnostics alongside outcome scores, and retaining contamination checks and
held-out evaluations.

\section*{Acknowledgment}
Ruoxi Jia and the ReDS lab acknowledge support from the National Science Foundation
through grants IIS-2312794, IIS-2313130, OAC-2239622, CNS-2424127, OAC-2613761, and
the Amazon-Virginia Tech Initiative for Efficient and Robust Machine Learning. We also
acknowledge Advanced Research Computing at Virginia Tech for providing computational
resources and technical support that have contributed to the results reported within this
paper.

\newpage
\bibliography{custom}

\clearpage

\appendix

\onecolumn
\begin{appendices}
\startcontents[appendices]
\printcontents[appendices]{}{1}{\setcounter{tocdepth}{3}}

\clearpage
\section{Extended Related Work}

\paragraph{Autonomous research and machine-learning agents.}
A growing line of work evaluates language-model agents in interactive environments where they read files, edit code, run commands, and observe feedback. AgentBench introduced a broad multi-environment benchmark for LLM agents~\citep{liu2023agentbench}; MLAgentBench and MLE-bench evaluate machine-learning experimentation and Kaggle-style engineering~\citep{huang2023mlagentbench,chan2024mle}; PostTrainBench evaluates whether CLI agents can post-train base LLMs under bounded compute~\citep{rank2026posttrainbench}; and Terminal-Bench broadens the setting to general command-line tasks~\citep{merrill2026terminal}. AutoResearch and AI-Scientist-style systems push further, placing agents in iterative loops of idea generation, code modification, experiment execution, and reporting~\citep{karpathy2026autoresearch,lu2024ai,yamada2025ai}. Conceptually, the AutoResearch framework extends to optimizing training data in the same way as for model and training configurations; preliminary discussion appears in \citet{karpathy2026autoresearch}, but a systematic study on the data front is absent. \bench{} fills that gap by isolating training-data curation as the central object of study.

\paragraph{Data-centric AI and data selection.}
Data-centric AI emphasizes dataset quality, coverage, and labeling as first-class levers for model improvement~\citep{ng2021data}. The literature spans data pruning, coreset selection, influence functions, data valuation, quality filtering, and training-dynamics metrics, etc.~\citep{koh2017understanding,ghorbani2019data,paul2021deep,sorscher2022beyond,northcutt2021confident}. EL2N and GraNd identify important examples early in training~\citep{paul2021deep}; instruction-tuning data selection has been studied via quality scoring, diversity, influence approximation, and targeted selection~\citep{xia2024less}; and Self-Instruct-style pipelines show that LLMs can also generate instruction data that requires its own filtering and auditing~\citep{wang2023self}. 
Recent works also include scaling laws components that allows efficiently optimizing large-scale training data (e.g., LLM pre-training) with experiments at much smaller scales~\citep{kang2023performance, kang2025autoscale, ye2025data}. Our work is motivated by this literature but targets a different question: a useful \emph{data agent} must not just execute a known method--it must decide which method family is appropriate, implement it under budget, inspect evidence, and revise the policy.

\paragraph{Data-science agents.}
Data-science benchmarks evaluate agents that analyze datasets, write code, produce notebooks, and build tabular models. DSEval covers the data-science lifecycle~\citep{zhang2024benchmarking}; DSBench draws realistic analysis and modeling tasks from sources such as Kaggle and ModelOff~\citep{jing2024dsbench}; and TML-Bench evaluates autonomous agents on tabular ML tasks under time limits~\citep{pinchuk2026tml}. These targets are valuable but are usually \emph{analysis} or \emph{predictive modeling}; the artifact being optimized is a model or report, not a training corpus. \bench{} instead measures whether agents can produce a \emph{data policy} whose value is measured by downstream model behavior on generative-model training.

\paragraph{Agent benchmarks.} Agent benchmarks such as AgentBench, MLAgentBench, MLE-bench, PostTrainBench, Terminal-Bench, AutoResearch, and AI-Scientist-style systems evaluate whether agents can operate in interactive environments, modify code, run experiments, and report results~\citep{liu2023agentbench,huang2023mlagentbench,chan2024mle,rank2026posttrainbench,merrill2026terminal,karpathy2026autoresearch,lu2024ai,yamada2025ai}. Data-centric AI and data-selection research provide many non-agentic methods for pruning, scoring, filtering, and valuing examples~\citep{ng2021data,koh2017understanding,ghorbani2019data,paul2021deep,sorscher2022beyond,northcutt2021confident,xia2024less,wang2023self}. Data-science-agent benchmarks evaluate analysis, notebook construction, and tabular modeling rather than the downstream value of a curated training corpus~\citep{zhang2024benchmarking,jing2024dsbench,pinchuk2026tml}. \bench{} fills the gap summarized in Table~\ref{tab:related}: it evaluates whether generalist agents can discover, implement, and revise data policies whose value is measured by downstream model behavior.

\begin{table}[t]
\centering
\small
\caption{\small{Positioning relative to agent and data-centric benchmarks. The table intentionally focuses on whether the benchmark makes training-data curation the central research object.}}
\label{tab:related}
\resizebox{0.95\textwidth}{!}{
\renewcommand{\arraystretch}{1.2}
\begin{tabular}{m{3.2cm}m{3cm}m{2.8cm}m{1.6cm}m{5cm}}
\toprule
\textbf{Work} & \textbf{Primary target} & \textbf{Agent actions} & \textbf{Data tasks?} & \textbf{Gap for autonomous data research} \\
\midrule
AutoResearch~\citep{karpathy2026autoresearch} & Training-recipe, code, and hyperparameter optimization & Modify code/configs, run experiments, report & No & Optimizes a pipeline under a fixed data distribution; \bench{} isolates the data distribution itself as the object of optimization. \\
AI Scientist~\citep{lu2024ai,yamada2025ai} & Automated research loops & Generate ideas, modify code, executes experiments, write paper & Partial & Data side is usually one of many levers, not isolated as the benchmark target. \\
MLAgentBench~\citep{huang2023mlagentbench} & ML experimentation & Train models, edit code, inspect output & Partial & Evaluates broad ML experimentation rather than data policy discovery. \\
MLE-bench~\citep{chan2024mle} & ML engineering competitions & Kaggle-style engineering & Partial & Data preparation matters, but objective is competition performance, not data curation mechanisms. \\
PostTrainBench~\citep{rank2026posttrainbench} & Autonomous LLM post-training & Fine-tune, curate data, evaluate & Partial & Evaluates end-to-end post-training; our focus is trace-level diagnosis and scaffolds for data strategy exploration. \\
\midrule
DSEval \citep{zhang2024benchmarking} / DSBench \citep{jing2024dsbench} / TML-Bench \citep{pinchuk2026tml} & Data science and tabular ML & Analyze data, produce notebooks, train tabular models & No & Data is an input to analysis/modeling rather than the artifact being optimized. \\
Data-centric AI / data selection~\citep{ng2021data,paul2021deep,sorscher2022beyond} & Dataset engineering & Usually non-agentic methods & Yes & Provides methods, but not an agent benchmark for autonomous data research. \\
\midrule
\bench{}\newline(\textit{this work}) & Training data curation & Select, mix, filter, rewrite, inspect, train, evaluate & Yes & Measures whether generalist agents can discover and execute data policies. \\
\bottomrule
\end{tabular} \renewcommand{\arraystretch}{1.0}}
\end{table}

\section{Details for \bench{}}\label{app:bench}

\subsection{Source data and models.}
The benchmark draws training pools from two public visual-instruction mixtures and fine-tunes one of four base VLMs.

\textbf{LLaVA-665K.}~Following~\citet{liu2024improved}, we use the \href{https://huggingface.co/datasets/liuhaotian/LLaVA-Instruct-150K/blob/main/llava_v1_5_mix665k.json}{\texttt{llava\_v1\_5\_mix665k.json}} mixture, materialized locally as a single Arrow dataset of $665{,}298$ rows with three columns (\texttt{images}, \texttt{texts}, \texttt{source\_subset}). Rows are tagged by their originating \texttt{source\_subset}, partitioning the pool into six groups: \texttt{coco} ($364{,}100$), \texttt{vg} ($86{,}417$), \texttt{ocr\_vqa} ($80{,}000$), \texttt{gqa} ($72{,}140$), \texttt{text\_only} ($40{,}688$, image-free ShareGPT turns), and \texttt{textvqa} ($21{,}953$).

\textbf{Vision-Flan.}~We additionally use the human-annotated Vision-Flan~\citep{xu2024vision} mixture, materialized as a single Arrow dataset with three columns (images, texts, source\_subset). It has over $191$ task types (approximately $1{,}000$ samples per task, following the released \texttt{annotation\_191-task\_1k} layout).

\textbf{Models.}~Each task pins a single \texttt{model\_key} so that fine-tuning always starts from the same pre-curation checkpoint:
\begin{itemize}
  \item \textbf{LLaVA-1.5-7B}~\citep{liu2024improved} -- visual instruction-tuning stage; we full-finetune the language model with the vision tower frozen. Checkpoint released with the artifact as \href{https://huggingface.co/anonneuripsmail/llava-1.5-7b-init}{\texttt{llava-1.5-7b-init}}.
  \item \textbf{Qwen2-VL-2B} (base)~\citep{wang2024qwen2} -- visual instruction-tuning stage; we full-finetune the entire model (vision + encoder + projector merger + language model). Checkpoint can be found in \href{https://huggingface.co/Qwen/Qwen2-VL-2B}{\texttt{Qwen2-VL-2B}}.
  
  \item \textbf{Qwen2.5-VL-3B-Instruct}~\citep{bai2025qwen25vltechnicalreport} -- post-training stage starting from the released instruction-tuned checkpoint (\href{https://huggingface.co/Qwen/Qwen2.5-VL-3B-Instruct}{\texttt{Qwen2.5-VL-3B-Instruct}}). We full-finetune the entire model (visual encoder + projector merger + language model), with no module frozen. 
  \item \textbf{SmolVLM-2.2B-Base}~\citep{marafioti2025smolvlm} -- visual
  instruction-tuning stage starting from the pre-instruct base checkpoint; we
  full-finetune the language model and the projector with the SigLIP vision tower
  frozen. Checkpoint can be found in
  \href{https://huggingface.co/HuggingFaceTB/SmolVLM-Base}{\texttt{SmolVLM-Base}}.
\end{itemize}

\textbf{DataComp-Small.}~For the CLIP pretraining instantiation we use the
\emph{Small} configuration of DataComp~\citep{gadre2023datacomp}, the
$12.8$M-pair filtering-track pool. After downloading we recovered
$8.77$M of the $12.8$M pairs ($68.5\%$ of the original pool, the
remainder lost to link rot) and resized all images to $256$ before
storage. The filtering-track rules fix the candidate pool, training
compute budget, model architecture, and evaluation suite; the agent's
only degree of freedom is the subset of pairs retained for training.

\subsection{Contamination audit}\label{app:contam}

Before fine-tuning, an agent may run a contamination audit on its candidate submission to check for overlap with the eight target benchmarks of \bench{} (HallusionBench, LLaVABench, MMBench, MMMU\_DEV\_VAL, MMStar, MMVet, MathVista\_MINI, OCRBench). We build a normalized question--answer string for each conversation turn in the submission and for each evaluation Q/A pair (lowercase, punctuation stripped, whitespace collapsed), and check (i)~exact matches via SHA-256 hashing and (ii)~near-duplicates via word $8$-gram overlap with similarity $\ge 0.8$. The audit aggregates per-benchmark exact-match rate $r_e$ and near-duplicate rate $r_n$ and returns \textsc{high\_risk} ($r_e \ge 0.10$ or $r_n \ge 0.20$), \textsc{warning} ($r_e \ge 0.05$ or $r_n \ge 0.10$), or \textsc{clean} otherwise. A \textsc{high\_risk} status forces the agent to revise its selection and re-audit before submission can proceed; \textsc{warning} does not block submission but reports the overlap to alert the agent; \textsc{clean} proceeds normally.

\subsection{Submission validation}

Submission is the gate that pins a candidate dataset as the input to the training stage. We enforce two conditions: (i)~\textbf{schema}: the path is loadable as a HuggingFace dataset with columns \texttt{images} and \texttt{texts}; (ii)~\textbf{budget}: the row count equals the task's \texttt{target\_rows}; A validated submission is archived into the run directory and any prior fine-tuning result for the iteration is invalidated, so submission acts as a commit point: the agent may iterate freely before calling submit, but once accepted the dataset is frozen for the rest of the iteration.

\subsection{Training}

\bench{} fixes the entire fine-tuning stage so that the only variable across an agent's iterations is the curated subset. Given a submitted subset, the harness fine-tunes the task's target model on a single NVIDIA A100 GPU for one epoch under a fixed recipe; no training hyperparameter is exposed to the agent. Each target model in the suite has its own recipe. For LLaVA-1.5-7B and SmolVLM-Base, the vision tower is frozen and only the language model and the multi-modal projector are updated; the detailed training configurations can be found in Table~\ref{tab:fixed-training-recipe} (LLaVA-1.5-7B, except for the Training hardware) and Table~\ref{tab:fixed-training-recipe-smolvlm} (SmolVLM-Base). For Qwen2-VL-2B and Qwen2.5-VL-3B-Instruct, the entire model is full-finetuned with FlashAttention-2 (Tables~\ref{tab:fixed-training-recipe-qwen2vl2b} and~\ref{tab:fixed-training-recipe-qwen25vl3b}).

\begin{table}[htbp]
\centering
\small
\caption{\small{Fixed fine-tuning recipe used for all agent and baseline submissions.}}
\label{tab:fixed-training-recipe}
\begin{tabular}{ll}
\toprule
\textbf{Component} & \textbf{Value} \\
\midrule
Initial model & LLaVA-1.5-7B init (\href{https://huggingface.co/anonneuripsmail/llava-1.5-7b-init}{\texttt{HF checkpoint}}) \\
Training hardware & 1 NVIDIA A100 GPU per model \\
Training epochs & 1 \\
Vision tower & Frozen \\
Per-device batch size & 1 \\
Gradient accumulation & 16 for one-GPU subset runs \\
Learning rate & $2\times 10^{-5}$ \\
Learning-rate schedule & Cosine \\
Warmup ratio & 0.03 \\
Precision & bfloat16 \\
Optimizer & Fused AdamW \\
Gradient checkpointing & Enabled \\
Formatting & LLaVA chat-template normalization \\
Loss masking & Assistant-token masking \\
\bottomrule
\end{tabular}
\end{table}

\begin{table}[htbp]
\centering
\small
\caption{\small{Fine-tuning settings for Qwen2-VL-2B in \bench{}.}}
\label{tab:fixed-training-recipe-qwen2vl2b}
\begin{tabular}{ll}
\toprule
\textbf{Component} & \textbf{Value} \\
\midrule
Initial model & Qwen2-VL-2B  \\
Training hardware & 1 NVIDIA A100 GPU \\
Training epochs & 1 \\
Trainable modules & Entire model \\
Per-device batch size & 1 \\
Gradient accumulation & 8 \\
Learning rate & $1\times 10^{-5}$ \\
Learning-rate schedule & Cosine \\
Warmup ratio & 0.03 \\
Weight decay & 0.1 \\
AdamW $\beta_2$ & 0.95 \\
Precision & bfloat16 \\
Optimizer & adamw\_torch\_fused \\
Attention implementation & FlashAttention-2 \\
Gradient checkpointing & Enabled \\
\bottomrule
\end{tabular}
\end{table}

\begin{table}[htbp]
\centering
\small
\caption{\small{Fine-tuning settings for Qwen2.5-VL-3B-Instruct in \bench{}.}}
\label{tab:fixed-training-recipe-qwen25vl3b}
\begin{tabular}{ll}
\toprule
\textbf{Component} & \textbf{Value} \\
\midrule
Initial model & Qwen2.5-VL-3B-Instruct \\
Training hardware & 1 NVIDIA A100 GPU \\
Training stage & Post-training  \\
Training epochs & 1 \\
Trainable modules & Entire model \\
Per-device batch size & 1 \\
Gradient accumulation & 8 \\
Learning rate & $1\times 10^{-5}$ \\
Learning-rate schedule & Cosine \\
Warmup ratio & 0.03 \\
Weight decay & 0.1 \\
AdamW $\beta_2$ & 0.95 \\
Precision & bfloat16  \\
Optimizer & adamw\_torch\_fused \\
Attention implementation & FlashAttention-2 \\
Gradient checkpointing & Enabled \\
\bottomrule
\end{tabular}
\end{table}

\begin{table}[htbp]
  \centering
  \small
  \caption{\small{Fine-tuning settings for SmolVLM-Base (2.2B) in \bench{}.}}
  \label{tab:fixed-training-recipe-smolvlm}
  \begin{tabular}{ll}
  \toprule
  \textbf{Component} & \textbf{Value} \\
  \midrule
  Initial model & SmolVLM-2.2B-Base \\
  Training hardware & 1 NVIDIA A100 GPU \\
  Training epochs & 1 \\
  Trainable modules & Language model + multi-modal projector (vision tower frozen) \\
  Per-device batch size & 16 \\
  Gradient accumulation & 1 \\
  Learning rate & $1\times 10^{-5}$ \\
  Learning-rate schedule & Cosine \\
  Warmup ratio & 0.03 \\
  Weight decay & 0.0 \\
  Precision & bfloat16 \\
  Optimizer & adamw\_torch\_fused \\
  Attention implementation & SDPA \\
  Gradient checkpointing & Enabled \\
  \bottomrule
  \end{tabular}
  \end{table}

For the DataComp instantiation, the harness follows the published
DataComp-Small filtering-track recipe verbatim
(\Cref{tab:fixed-training-recipe-datacomp}): a CLIP \mbox{ViT-B/32} is
trained from scratch on the agent's filtered subset using
OpenCLIP, with total training compute fixed at $12.8$M processed
image--text pairs regardless of the size of the filtered subset, so a
full-pool run is ${\sim}1.5$ passes over the $8.77$M available pairs
(via the framework's with-replacement WebDataset sampling). All
hyperparameters are the Small-track defaults reported by
\citet{gadre2023datacomp} and instantiated in the
\href{https://github.com/mlfoundations/datacomp}{\texttt{mlfoundations/datacomp}}
repository; no training hyperparameter is exposed to the agent.

\begin{table}[htbp]
\centering
\small
\caption{\small{CLIP pretraining settings for the DataComp instantiation
(Small filtering track). All values are the published Small-track
defaults from~\citet{gadre2023datacomp} (DataComp \texttt{scale\_configs.py})
and OpenCLIP defaults inherited through DataComp's \texttt{train.py}
driver. The only setup choice particular to our hardware is running the
global batch on a single H100; the DataComp default global batch size
is also $4096$, so no accumulation or per-GPU rescaling is required.}}
\label{tab:fixed-training-recipe-datacomp}
\begin{tabular}{ll}
\toprule
\textbf{Component} & \textbf{Value} \\
\midrule
Initial model & CLIP ViT-B/32 (random init) \\
Training framework & OpenCLIP (DataComp \texttt{train.py} driver) \\
Training hardware & 1 NVIDIA H100 GPU \\
Image storage resolution & $256\times 256$ (pre-downloaded) \\
Model input resolution & $224\times 224$ (OpenCLIP ViT-B/32) \\
Total samples seen & $12{,}800{,}000$ image--text pairs (Small-track budget) \\
Data sampling & WebDataset, with replacement (\texttt{--dataset-resampled}) \\
Global batch size & $4096$ \\
Optimizer & AdamW \\
Learning rate & $5\times 10^{-4}$ \\
LR schedule & Cosine decay \\
Warmup steps & $500$ \\
Weight decay & $0.2$ \\
AdamW $(\beta_1, \beta_2)$ & $(0.9,\ 0.98)$ \\
AdamW $\epsilon$ & $1\times 10^{-6}$ \\
Precision & Automatic mixed precision (fp16) \\
Gradient checkpointing & Enabled \\
Loss & Local-loss with \texttt{--gather-with-grad} \\
Wall-clock per run & ${\sim}2$ hours training + ${\sim}2$ hours evaluation\\
\bottomrule
\end{tabular}
\end{table}

\subsection{Evaluation}

 The benchmark fixes the evaluation stage and exposes no evaluation
 parameter to the agent.  The protocol -- VLMEvalKit wrapper, the eight
 target benchmarks, the shared Qwen3.5-27B judge endpoint (\Cref{tab:judge-server-config}), and the aggregation rule (\Cref{tab:benchmark-normalization}) -- is exactly the one described in \Cref{appendix:eval}.

For the DataComp instantiation, the harness uses the standard
DataComp evaluation suite~\citep{gadre2023datacomp}: $38$ zero-shot
tasks spanning ImageNet and its distribution shifts, VTAB
classification tasks, retrieval (Flickr30k, MSCOCO), and additional
classification benchmarks. We report the unweighted mean accuracy
across the $38$ tasks, matching the filtering-track leaderboard.
Evaluation takes ${\sim}2$ hours per run on a single H100.

\section{Scaffolds (open-prompt, light, heavy)}\label{app:scaf}
\subsection{Common Failures for Open-Prompt Agents}\label{sapp:fails1}

Table~\ref{tab:failures} summarizes common failure modes of open-prompt generalist agents in data curation tasks.
Table~\ref{tab:taxonomy} provides a non-exhaustive taxonomy of data-curation strategy families. The families are partially disconnected: moving from one to another often requires new tools, proxies, and assumptions.

\begin{table}[t]
\centering
\small
\caption{\small{Common failure modes of open-prompt generalist agents on data curation tasks.}}
\label{tab:failures}\vspace{0.5em}
\resizebox{0.95\textwidth}{!}{
\renewcommand{\arraystretch}{1.1}
\begin{tabular}{p{3.2cm}p{6cm}p{5cm}}
\toprule
\textbf{Failure mode} & \textbf{Description} & \textbf{Potential scaffold} \\
\midrule
Only doing heuristics & Agent repeatedly applies length, metadata, or score filters with weak justification. & Require policy-family novelty or evidence-backed ablations. \\
Vibe optimization & Agent states that data is ``higher quality'' without measurement. & Data audit reports, quality classifiers, and hypothesis templates. \\
Optimizing random seeds & Agent attributes noisy gains to data policy or spends budget on seeds. & Fixed seed protocols and paired comparisons. \\
Not using provided skills & Agent receives a skill list but never calls or implements the skills. & Skill cards with executable entry points and mandatory decision checkpoints. \\
No original research & Agent repeats known defaults and does not propose new data mechanisms. & Force comparison across policy families before local search. \\
Shallow eval-log reading & Agent reads aggregate score but not raw failures. & Structured error extraction, clustering, and traceable rewrite plans. \\
Overfitting visible evals & Agent creates examples too close to evaluation outputs. & Contamination checks, hidden evals, and red-team audits. \\
Heavy scaffold contradiction & Detailed pipelines improve scores but reduce autonomy. & Report scaffold strength as an experimental variable. \\
 \textit{(rewriting)} Selection without rewriting & Agent is told to rewrite but only selects existing examples. & Separate tools and budgets for rewrite generation and validation. \\
\bottomrule
\end{tabular}\renewcommand{\arraystretch}{1.0}}
\end{table}

\begin{table}[t]
\centering
\caption{\small{A non-exhaustive taxonomy of data-curation strategy families. The families are partially disconnected: moving from one to another often requires new tools, proxies, and assumptions \citep{bhardwaj2024machine,maghoumi2024nemocurator,nvidia2024nemomicroservices}.}}\vspace{0.5em}
\label{tab:taxonomy}
\small
\resizebox{0.9\textwidth}{!}{
\renewcommand{\arraystretch}{1.1}\begin{tabular}{p{0.22\linewidth}p{0.35\linewidth}p{0.33\linewidth}}
\toprule
\textbf{Strategy family} & \textbf{Example interventions} & \textbf{Required machinery} \\
\midrule
Quality and format & Remove malformed, low-quality, or unsafe examples & Validators, quality models, metadata \\
Distribution/domain & Tune source and domain mixture ratios & Domain labels, mixture search \\
Semantic coverage & Cluster examples, maximize diversity, cover skills & Embeddings, clustering, facility-location objectives \\
Task/instruction level & Balance task types, tools, reasoning modes & Task classifiers, taxonomies \\

Training dynamics & Select by loss, EL2N, gradient norm, forgetting & Warm-up training, per-example statistics \\
Dedup/contamination & Remove near-duplicates and benchmark leakage & MinHash, embedding similarity, benchmark filters \\
Style and response form  \newline(\textit{rewriting})& Rewrite for concise, tool-using, or stepwise style & Rewriting model, validators \\
Generation feedback \newline(\textit{data synthesis}) & Add data from model failures or self-play & Eval logs, error classifiers, synthetic generation \\
\bottomrule
\end{tabular}\renewcommand{\arraystretch}{1.0}}
\end{table}

\subsection{Light Scaffolds}\label{app:light-scaffold}

The design principle of our light scaffolds is to remain task-agnostic so that the same prompt could plausibly ship as a default for any data-curation task. The goal of this paper is not to add another data-curation strategy but to make ready-to-use generalist agents practical for frontline practitioners. We explore two strategies that best embody that ideology: (I) providing a list of curation strategies organized by policy family and instructing the agent to explore them, and (II) compiling research papers on data tasks into skill cards for the agent to reference.

\paragraph{I. Concise instructions toward desired trajectories.}
To counter the agent's tendency to fixate on a single strategy family, we surface a list of diverse operations upfront and prompt the agent to explore beyond them. The instruction is deliberately framed at the level of policy families rather than task-specific recipes: this expands the agent's awareness of the solution space without prescribing implementations and preserves autonomy over \emph{how} each idea is instantiated.

\paragraph{II. Research papers as skill cards.}
The value of human data experts lies in knowing which methods are viable for a given task and when each is most likely to help. Rather than hand-injecting that expertise per task, we compile relevant research papers into reusable skill cards that the agent can consult on its own. Concretely, we prompt Claude Code with an automated pipeline (Appendix~\ref{app:prompts}) to compile a library of skill cards from VLM data-curation papers, each formatted as a decision-first cookbook entry rather than a paper summary. The compilation is offline--skill cards are prepared before any agent run--and the agent prompt is augmented with two short notes: list all available skill cards at the start of each session, and \emph{use the reference skills when helpful}.

\begin{tcolorbox}[enhanced, width=\linewidth,
  title=\small{\textbf{Skill cards} compiled from research papers on \textit{data curation vision-language models (VLMs)}},
  colback=lightgray!10!white, colframe=black]
\small
\#\# Characteristics of the skills\\

-- Four top-level buckets, each a stage of the VLM data pipeline: data-acquisition, data-curation, data-selection, data-synthesis\\
-- Each skill is one folder containing exactly one file: `SKILL.md` with a title that combines the paper's short title and source.\\
-- 1 paper -> 1 skill. No multi-paper aggregations; no extra assets (no code, no figures, no PDFs).\\

**Per-file structure (rigidly templated)**\\

**Tone / philosophy**\\
-- Decision-first ("Use this skill when…" / "Avoid it when…") rather than paper summary.\\
-- Procedures are skeletons — the *shape* of the method, not full reproduction. Each step ends with "See paper for details."\\
-- Knowledge-cards optimized for a downstream agent to retrieve and act on, not for human reading. They look like cookbook recipe cards keyed by paper.
\normalsize
\end{tcolorbox}

\paragraph{Common Failures under Light Scaffolding.}\label{asec:light-fail}
Table~\ref{tab:prompt-failures} summarizes the failures: light interventions improve the written plan but rarely the executed trajectory. The agent acknowledges the broader options and then chooses the easiest implementable heuristic. The bottleneck is not knowledge--the agent often \emph{names} training dynamics, semantic clustering, or rewriting--but the conversion of a high-level idea into a valid, budget-aware implementation with the right diagnostics. This is the recurrent pattern: current generalist agents are strong executors but weak autonomous researchers, capable of running the benchmark, writing scripts, debugging failures, and improving a local policy, yet unable to operationalize abstract data-research ideas without additional structure.

\begin{table}[h]
\centering
\small
\caption{\small{Interventions with light-scaffolding and observed failure modes.}}\vspace{0.5em}
\label{tab:prompt-failures}
\resizebox{0.95\textwidth}{!}{
\renewcommand{\arraystretch}{1.1}\begin{tabular}{p{3.5cm}p{4.5cm}p{5.6cm}}
\toprule
\textbf{Intervention} & \textbf{Intended effect} & \textbf{Observed behavior} \\
\midrule
Standard instruction & Let the agent autonomously improve data. & Defaults to filtering, deduplication, simple score thresholds. \\
Ask for broad solutions & Encourage movement across policy families. & Plans mention many families; implementation remains local. \\
Ask to avoid heuristics & Suppress trivial length/score filters. & Agent renames heuristics as ``quality criteria'' or uses them indirectly. \\
Provide flat skill list & Expose many approaches. & Skills are rarely used or used superficially without correct evidence. \\
Ask to inspect raw eval logs & Encourage targeted rewriting. & Agent reads summary metrics or selects examples, but rarely rewrites with grounded error taxonomy. \\
\bottomrule
\end{tabular}\renewcommand{\arraystretch}{1.0}}
\end{table}

\subsection{Heavy Scaffolds}\label{app:heavy-scaffold}

\paragraph{I. Self-Research} reshapes the workflow into a structured research loop where every experiment must be motivated by a concrete observation, documented with a hypothesis, and logged with its findings--positive or negative--to inform the next iteration. Examples of agent-written research plans are provided in the code repo.

\begin{center}
\begin{tcolorbox}[enhanced, width=0.95\linewidth,
  title=\small{\textbf{Heavy-scaffolding I: }Self-research},
  colback=lightgray!10!white, colframe=black]
\small
\begin{itemize}[leftmargin=0.5em, itemsep=1pt]
    \item \textit{Research-first mandate} requiring that every iteration (beyond baselines) begin with actual data research rather than intuitive code changes.  Before implementing the curation script, the agent is instructed to first write a research.md for the iteration containing an observation, a hypothesis, and a minimal change — all grounded in concrete data findings, not generic intuition. 
    
\item \textit{Blocklist of exhausted strategies:} The research version adds an explicit list of patterns that are "out" — strategies based on trivial operations and shallow ideas. 

\item \textit{Richer result logging schema:} The agent is instructed to document outcomes at each iteration with an explicit template including three additional fields--"skill ref" (which prior research.md commit this iteration builds on), "failure mode" (what new finding the run revealed, tied to specific benchmarks), and "next skill candidate" (the next research question to investigate). This creates a chain of reasoning across iterations to discourage ungrounded moves without a reasonable rationale or logic. 
\end{itemize}
\normalsize
\end{tcolorbox} 
\end{center}

\paragraph{II. Adapt Research Papers} hardens the soft ``prioritize skills'' guidance from light-scaffolding II into a strict protocol: every iteration must cite a specific skill, adapt a concrete procedure from it, rotate across skill categories, and avoid a defined set of low-effort heuristics. The result is a more disciplined and auditable research loop.

\begin{center}
\begin{tcolorbox}[enhanced, width=0.95\linewidth,
  title=\small{\textbf{Heavy-scaffolding II: }Adapt Research Papers},
  colback=lightgray!10!white, colframe=black]
\small
\begin{itemize}[leftmargin=0.5em, itemsep=1pt]
    \item \textit{Session initialization}: listing skills -> deeply reading skills: at the start of the session, the agent must browse all skill categories, pick at least 3 skills per category that seem relevant, and read those skill cards in full before writing its curation script.  The agent is also prompted to use `Agent`/`Explore` tools to scan quickly through the one-line decision notes in the skill cards. The pipeline raises the bar from awareness to comprehension.

\item \textit{Strategy must be skill-grounded}: formalizing the requirement that the curation script at each iteration must be grounded in a specific skill card. The workflow is: identify a failure mode from the last result, pick 1–3 relevant skills, read them fully, then state in the logs a specific skill card and a one-line adaptation.  This changes implicit/soft guidance in the light-scaffolding ("prioritize exploration with reference skills") to a hard rule.

\item \textit{Category rotation requirement}: the heavy-scaffolding requires that before repeating a skill category, the agent must attempt at least one iteration from each of the other three. The `next skill candidate` must also come from a category not used in the last three iterations unless justified. 

\item \textit{Blocklist of exhausted strategies} and \textit{Richer result logging schema} are enforced similarly to Heavy-scaffolding I. The agent is forbidden from citing a skill as mere "inspiration" rather than an adapted procedure. Not citing a specific skill card in a non-baseline row is explicitly called a "rule violation."

\end{itemize}
\normalsize
\end{tcolorbox} 
\end{center}

\subsection{Rewriting Scaffolds}

\paragraph{Scaffolding for rewriting.}
The rewriting scaffold introduces two capabilities absent from the selection-only scaffolds. First, \emph{eval-log reading}: after each evaluation, the agent samples 5--15 incorrect predictions per benchmark, categorizes the failure mode (e.g., OCR misread, format violation, hallucinated option, missing reasoning step), and cross-references the training data to identify which examples--or gaps--produced the failure. Second, \emph{targeted rewriting}: the agent constructs a rewrite recipe specifying (a) which slice of the training data to target, (b) what transformation to apply via the MLLM, and (c) the expected effect on the identified failure mode. The recipe is logged alongside the iteration result, creating an auditable chain from failure observation to rewrite action to training outcome. Full scaffold details are in Appendix~\ref{app:prompts}.

We adopt the self-research heavy scaffold for rewriting. Each iteration must begin with evidence from evaluation logs or data inspection, name a concrete failure mode, select source examples tied to that failure, rewrite them with an external MLLM tool, and validate the rewritten outputs before training. The scaffold therefore changes the task from ``pick the best examples'' to ``pick and improve examples under a fixed training budget.'' Compared with template rewriting, the setup is more agentic: the rewrite policy is not fixed in advance, and the agent must decide which examples to rewrite, what transformation to apply, and how aggressively to use the rewriting model.

\section{Experimental Details and Additional Results}
\label{app:exps}
\subsection{Terminal Agents}
\label{sapp:agents}

All agents interact with the \bench{} environment through a terminal
interface: they read files, write and edit scripts, run shell commands,
and observe stdout/stderr. The agents differ in their backbone model,
harness, and hosting, but share the same workspace, CLI, and
submission gate.

\paragraph{Claude Code.}
We use the Claude Code terminal agent~\citep{anthropic2025claudecode}
with two backbone models: Opus~4.7 (primary) and Sonnet~4.6 (ablation).
Claude Code provides built-in tool use for file editing, command
execution, and multi-step planning. No additional plugins or custom
tools are added beyond the benchmark CLI.

\paragraph{Codex.}
We use OpenAI's Codex terminal agent with two backbone models:
GPT-5.4 (primary) and GPT-5.3 (ablation). Codex operates in a
sandboxed environment with the same file-system layout and CLI
access as the Claude Code runs.

\paragraph{Open-source models via OpenHands.}
To evaluate open-source agent backbones, we use the OpenHands
harness~\citep{wang2025openhands}, which wraps an LLM in a
terminal-agent loop with file editing, shell execution, and
observation parsing. We evaluate two models: Kimi~K2.5 and
Qwen3.5-397B. Both are served via the Together AI
API\footnote{\url{https://www.together.ai/}} using an
OpenAI-compatible chat-completion endpoint, which OpenHands
consumes as its backend LLM. Inference parameters (temperature,
max tokens) follow the OpenHands defaults for each model.

\paragraph{Shared configuration.}
Across all agents, the Docker workspace, candidate-pool access,
benchmark CLI, submission validation, contamination audit, training
recipe, and evaluation suite are identical. The only differences
are the agent backbone and harness; no agent receives
task-specific tools or data-curation expertise beyond what the
scaffold condition specifies (\Cref{tab:prompt-excerpts}).

\subsection{DataComp Filtering Setup}
\label{sapp:datacomp-setup}

For the CLIP pretraining instantiation, all agent and baseline submissions
use the DataComp-Small filtering-track configuration described in
\Cref{app:bench}: the same $12.8$M-pair candidate pool
(of which $8.77$M pairs were recoverable after link rot, see
\Cref{app:bench}), the same OpenCLIP ViT-B/32 training recipe
(\Cref{tab:fixed-training-recipe-datacomp}), and the same $38$-task
zero-shot evaluation suite~\citep{gadre2023datacomp}. This subsection
records only the differences in \emph{what was run} rather than \emph{how
it was run}.

\paragraph{Agent}
We evaluate Claude Code (Opus 4.7) on this instantiation. The agent
operates over the metadata fields available in the DataComp pool,
implements a filtering policy, submits the filtered subset through the
harness CLI, and revises after observing the DataComp-suite score.

\paragraph{Baseline}
We compare against the published \emph{top-$30\%$ CLIP score}
DataComp filtering baseline~\citep{gadre2023datacomp}: pairs are kept
when their similarity score under a frozen CLIP ViT-L/14 reference
encoder is in the top $30\%$ of the pool, after which a ViT-B/32 is
trained from scratch under the same Small-track recipe.

\paragraph{Resources per session}
A single training-and-evaluation cycle for one filtered subset takes
${\sim}4$ hours of H100 time (${\sim}2$~h training, ${\sim}2$~h
evaluation, \Cref{tab:fixed-training-recipe-datacomp}).

\subsection{Shared Training and Evaluation Protocol}
\label{sapp:train-eval-protocol}

In experiments with fine-tuning pre-trained LLaVA-1.5-7B models on examples from the LLaVA-665K dataset, all submitted
datasets are trained and evaluated through the same wrapper.  Each submitted
subset is fine-tuned for one epoch with the fixed recipe in
\Cref{tab:fixed-training-recipe} and evaluated by the same eight-benchmark
VLMEvalKit pipeline~\citep{duan2024vlmevalkit}.

\paragraph{Source data and model.}
All subsets are drawn from the LLaVA-665K mixture~\citep{liu2024improved}
released on Hugging Face at \url{https://huggingface.co/datasets/liuhaotian/LLaVA-Instruct-150K/blob/main/llava_v1_5_mix665k.json},
materialized locally as five Arrow splits: \texttt{coco} (364,100),
\texttt{gqa} (72,140), \texttt{ocr\_vqa} (80,000), \texttt{textvqa}
(21,953), \texttt{vg} (86,417), and \texttt{text\_only} (40,688). Training always starts from the same
pre-data-curation LLaVA-1.5-7B checkpoint~\citep{liu2024improved},
released with the artifact as
\href{https://huggingface.co/anonneuripsmail/llava-1.5-7b-init}{\texttt{HF checkpoint}}.

\paragraph{Training.}
The per-model procedure is the same across baseline families, budgets, and
agent submissions.  Each submitted dataset is fine-tuned on one NVIDIA A100 GPU
from the same LLaVA-1.5-7B initialization for one epoch with the vision tower
frozen.  Training uses PyTorch~\citep{paszke2019pytorch} with Accelerate
support~\citep{gugger2022accelerate}.  The trainer uses the same LLaVA
chat-template normalization and assistant-token masking for all subsets.
\Cref{tab:fixed-training-recipe} summarizes the fixed recipe.

\paragraph{Evaluation.}
\label{appendix:eval}
Each trained checkpoint is evaluated through the repository's VLMEvalKit wrapper
rather than by manually invoking benchmark scripts.  The wrapper isolates each
benchmark's artifacts, archives stale outputs before reruns, and refuses
eval-only mode if the expected trained checkpoint is missing.  All checkpoints
are evaluated on the same eight VLMEvalKit benchmarks: MMVet~\citep{yu2023mm},
LLaVA-Bench~\citep{liu2023visual}, MMBench~\citep{liu2024mmbench}, MMMU
validation~\citep{yue2024mmmu}, MMStar~\citep{chen2024we},
MathVista-Mini~\citep{lu2023mathvista}, OCRBench~\citep{liu2024ocrbench}, and
HallusionBench~\citep{guan2024hallusionbench}.  Candidate-model inference is
run by VLMEvalKit~\citep{duan2024vlmevalkit} with
vLLM~\citep{kwon2023efficient}.  Judge-based metrics use a separate
OpenAI-compatible Qwen3.5-27B endpoint~\citep{team2026qwen3} for all baseline
runs.  The server-side judge configuration is summarized in
\Cref{tab:judge-server-config}.

\begin{table}[htbp]
\centering
\footnotesize
\caption{\small{Judge-server configuration for judge-based VLMEvalKit metrics.
This endpoint is separate from candidate-model evaluation inference.}\normalsize}
\label{tab:judge-server-config}
\begin{tabular}{p{4cm}p{5cm}}
\toprule
\textbf{Setting} & \textbf{Value} \\
\midrule
GPU type & 2 NVIDIA RTX A6000 GPUs \\
Model & \texttt{Qwen/Qwen3.5-27B} \\
Tensor parallel size & 2 \\
GPU memory utilization & 0.90 \\
Maximum model length & 2048 tokens \\
Maximum concurrent sequences & 64 \\
Precision & bfloat16 \\
Reasoning parser & \texttt{qwen3} \\
Thinking mode & Disabled via \texttt{enable\_thinking=false} \\
Request defaults & VLMEvalKit/OpenAI-wrapper defaults unless a benchmark wrapper overrides them: \texttt{temperature=0}, \texttt{max\_tokens=2048}, \texttt{n=1}. \\
\bottomrule
\end{tabular}
\end{table}

\paragraph{Aggregation and reporting.}
For each trained model, the aggregate score is the mean of the eight benchmark
values after applying the transformations in
\Cref{tab:benchmark-normalization}.  These transformations put all components
on an approximate 0--100, higher-is-better scale.  The resulting average is the
run-level score used for reporting.

\begin{table}[htbp]
\centering
\small
\caption{\small{Benchmark normalization used to compute the aggregate score.}}
\label{tab:benchmark-normalization}
\setlength{\tabcolsep}{5pt}
\begin{tabular}{p{0.24\linewidth}p{0.34\linewidth}p{0.30\linewidth}}
\toprule
\textbf{Benchmark} & \textbf{Raw value} & \textbf{Value used in aggregate} \\
\midrule
\texttt{MMVet} & Score reported by VLMEvalKit & Unchanged \\
\texttt{LLaVA-Bench} & Score reported by VLMEvalKit & Unchanged \\
\texttt{MMBench} & Accuracy in $[0,1]$ & $100 \times$ accuracy \\
\texttt{MMMU} validation & Accuracy in $[0,1]$ & $100 \times$ accuracy \\
\texttt{MMStar} & Accuracy in $[0,1]$ & $100 \times$ accuracy \\
\texttt{MathVista-Mini} & Score reported by VLMEvalKit & Unchanged \\
\texttt{OCRBench} & Score out of 1000 & Score $/10$ \\
\texttt{HallusionBench} & aAcc, fAcc, and qAcc & Mean of aAcc, fAcc, and qAcc \\
\bottomrule
\end{tabular}
\end{table}


\subsection{Non-Agent Baseline Construction}
\label{sapp:baseline-protocol}

This subsection documents the non-agent baselines for fine-tuning pre-trained LLaVA-1.5-7B models on examples from the LLaVA-665K dataset: simple sampling references,
human-designed selection baselines, and a rewriting baseline.  All baselines
are trained and evaluated with the shared protocol in
\Cref{sapp:train-eval-protocol}.

\paragraph{Baseline families.}
\Cref{tab:baseline-families} summarizes the non-agent references.  ICONS and
ARDS are used as released selected-example pools: we randomly sample the
requested budget from each pool.  Template Matters first samples from LLaVA-665K dataset
and then applies a programmatic template rewrite; this baseline does not call an
LLM.

\begin{table}[htbp]
\centering
\footnotesize
\caption{\small{Non-agent baseline families.}}
\label{tab:baseline-families}
\setlength{\tabcolsep}{4pt}
\begin{tabular}{p{0.20\linewidth}p{0.46\linewidth}p{0.25\linewidth}}
\toprule
\textbf{Baseline} & \textbf{Construction} & \textbf{Reference / artifact} \\
\midrule
Random & Sample $B$ examples from LLaVA-665K & \href{https://huggingface.co/datasets/liuhaotian/LLaVA-Instruct-150K/blob/main/llava_v1_5_mix665k.json}{HF data} \\
ICONS & Sample $B$ examples from the released ICONS pool & \citep{wu2024icons}; \href{https://github.com/princetonvisualai/icons}{GitHub} \\
ARDS & Sample $B$ examples from the released ARDS pool & \citep{yang2025data}; \href{https://github.com/xyang583/ARDS}{GitHub} \\
Template Matters & Sample from LLaVA-665K, then apply template rewrite & \citep{wang2024template}; \href{https://github.com/shijian2001/TemplateMatters}{GitHub} \\
\bottomrule
\end{tabular}
\end{table}

\paragraph{Budgets and replicates.}
Random and Template Matters are evaluated at $10\mathrm{k}$, $20\mathrm{k}$,
$50\mathrm{k}$, $100\mathrm{k}$, and $200\mathrm{k}$ examples.  ICONS and ARDS
are evaluated at $10\mathrm{k}$, $20\mathrm{k}$, $50\mathrm{k}$, and
$100\mathrm{k}$ examples.  For each baseline family and subset size, we
generate up to six independently seeded runs.  Within each run, all
baseline-family generators share the same run-level seed, so method comparisons
are paired when the corresponding method-size cells are available.  Different
runs use different seeds.  We report the mean and sample standard deviation
across available run-level scores.  If a method-size cell has fewer than six
completed runs, we report the available count rather than treating the cell as
complete.


\subsection{Additional Results and Ablation Studies}\label{sapp:ablations}

We run four additional ablations along the axes of \bench{}'s task specification and the agent setup. Each
ablation primarily varies one axis while holding the rest fixed at the original open-prompt setting
(LLaVA-665K $\to$ LLaVA-1.5-7B, $B = 10$k, default scaffold, Claude Code Opus 4.7 backbone).
The budget and scaffold ablations additionally replicate across both Claude and Codex
backbones to test whether their effects hold across model families. 

\textbf{Selection budget.}
We vary $B \in \{10\text{k}, 20\text{k}, 50\text{k}\}$ on fine-tuning pre-trained LLaVA-1.5-7B models on examples from the LLaVA-665K dataset curated by Claude Code (Opus 4.7)/Codex (GPT-5.4). This tests whether the agent's gain over a random
subset scales with budget.
\textbf{Task generalization.}
We vary $M_0$ and $\dataset_{\mathrm{pool}}$ at $B = 10$k under Claude Code Opus 4.7: two
alternative target models—\textit{Qwen2-VL-2B} (base-model fine-tuning) and
\textit{Qwen2.5-VL-3B} (post-training)—and one alternative data pool, \textit{Vision-Flan}.
This tests whether the curation strategies the agent develops transfer across model scales,
training regimes, and pool structures, or are tuned to LLaVA-665K's specific subset
organization.
\textbf{Agent backbone.}
We compare four agent backbones at $B = 10$k under the default scaffold: \textit{Claude Code with
Opus 4.7}, \textit{Claude Code with Sonnet 4.6}, \textit{Codex with GPT-5.4}, and \textit{Codex with
GPT-5.3}. This tests how different agent backbones perform on the same task and scaffold.
\textbf{Scaffold.}
We compare three scaffolds at $B = 10$k under both Claude and Codex: \textit{open-prompt} (minimal-scaffolded), \textit{Heavy Scaffold I} (self-research), and \textit{Heavy Scaffold II} (adapt papers).

Table~\ref{tab:codex-heavy} shows results for interventions with heavy-scaffolding (Codex). Both heavy scaffolding substantially increase grounded iterations while only Heavy Scaffold II (Adapt papers) significantly increases iterations with new-policy-family moves and eliminates shallow operations, which also leads to visibly improved outcomes.

\begin{table*}[h]
\centering
\caption{\small{Interventions with heavy-scaffolding (Codex). Autonomous data curation experiments: each row is one commit produced by the agent. ``Keep'' rows improved the composite accuracy; ``discard'' rows were rolled back. Reporting avg. scores and “\% of full gain” (\% of the improvement from the base model to full-data fine-tuning). The four right-most columns rate each iteration on whether it introduces a new policy family, is grounded in evidence, is effective (improves accuracy), or is trivial/heuristic. {\large$\circ$} = yes, {\small\ding{55}} = no. \textcolor{Green}{Green} numbers show improvement over the open-prompt baseline. Best accuracy and quality scores marked in \textbf{bold}}}
\label{tab:codex-heavy}
\resizebox{0.9\textwidth}{!}{%
\renewcommand{\arraystretch}{1.1}\begin{tabular}{p{2.5cm}p{2cm} p{4cm} C{1.8cm}C{1.5cm}C{1.5cm}C{1.5cm}|c}
\toprule
\textbf{Intervention} & \textbf{Avg. score/\%} & \textbf{Description} & \textbf{New Policy} & \textbf{Grounded} & \textbf{Effective} & \textit{\text{Shallow}} & \textbf{Execution}\\
\midrule
\textit{Open-prompt}\newline(\textit{baseline, 10k}) & \textit{33.53 / 57\%} & OCR text-rich ratio 60/40  & \textit{3/10} & \textit{4/10} & \textit{\textbf{4/10}} & \textit{6/10} & \textit{10/10}\\
\midrule
Self-research\newline(Heavy I, 10k)& 32.96 & soften region bonus in weighted sampler & 3/10 & \textbf{10/10}\newline\textbf{\textcolor{Green}{(+6/10)}} & 3/10 & 6/10 & 10/10\\
\textbf{Adapt papers}\newline (Heavy II, 10k) & \textbf{34.28 / 66\%\newline \textcolor{Green}{(+0.75/+9\%)}} & ALIGN/RedCaps-style minimal filtering with community-supported noisy pool retention & \quad\textbf{9/10}\newline\textbf{\textcolor{Green}{(+6/10)}} & \,\,\,\textbf{10/10}\newline\textbf{\textcolor{Green}{(+6/10)}} & 3/10 & \quad\textbf{1/10}\newline\textbf{\textcolor{Green}{(-5/10)}} & 10/10\\
\midrule
Random 100k & 33.7\tiny{$\pm$0.2}\normalsize/59\%\tiny{$\pm$3\%}\normalsize & - & - & - & - & - & -\\
Random 200k & 34.0\tiny{$\pm$0.2}\normalsize/63\%\tiny{$\pm$3\%}\normalsize & - & - & - & - & - & -\\
ICONS 100k \citep{wu2024icons}  & 34.5 / 69\% & - & - & - & - & - & -\\
ARDS 100k \citep{yang2025data}  & 34.1 / 64\% & - & - & - & - & - & -\\
\bottomrule
\end{tabular}\renewcommand{\arraystretch}{1.0}
}
\end{table*}

Table~\ref{tab:ablation-backbone} shows results for ablation studies with \textbf{different agent backbone} on the multimodal instruction-tuning data-selection task. Different agent backbones moderately affect the final outcome without changing the general patterns in comparison with the random baseline or human solutions.

\begin{table}[h]
\centering
\small
\caption{\small{
Results for ablation studies with \textbf{different agent backbone} on the multimodal instruction-tuning data-selection task. Scores averaged over 8 benchmarks. \textcolor{Green}{Green} numbers show improvement over the random 10k baseline. ``\% of full gain'' column measures the fraction of the improvement from the base model to full-data fine-tuning that is achieved by each setting. Best outcomes are marked in \textbf{bold}. Different agent backbones moderately affect the final outcome without changing the general patterns in comparison with 
the random baseline or human solutions.}}\vspace{0.5em}
\label{tab:ablation-backbone}
\resizebox{0.9\textwidth}{!}{
\begin{tabular}{l|l|lll|l}
\toprule
\textbf{Setting} & \textbf{Data size} & \textbf{Avg. score} & \textbf{$\Delta$ over base} & \textbf{\% of full gain} & \textbf{Notes} \\
\midrule
Base model & 0 & 28.8 & 0.0 & 0\% & pre-trained \\
Full-data fine-tuning & 665k & 37.1 & +8.3 & 100\% & official checkpoint \\
\midrule
\textit{Random 10k (baseline)}& \textit{10k} & \textit{32.5} & \textit{+3.7} & \textit{45\%} & \textit{Best of 10 runs}\\
\midrule
ICONS 10k  & 10k & 33.3 & +4.5 & 54\% & \citet{wu2024icons} \\
ARDS 10k & 10k & 33.2 & +4.3 & 52\% & \citet{yang2025data} \\
\midrule
Agent (open-prompt)  & 10k & \textbf{34.2  \textcolor{Green}{(+1.7)}} & +5.3 & \textbf{65\% \textcolor{Green}{(+20\%)}} & Claude Code (Sonnet 4.6), 10 iter.\\
Agent (open-prompt)  & 10k & 33.3 \textcolor{Green}{(+0.8)} & +4.5 & 54\% \textcolor{Green}{(+9\%)}& Codex (GPT-5.3), 10 iterations\\
\midrule
Random 50k & 50k & 33.4\tiny{$\pm$0.2} & +4.6\tiny{$\pm$0.2} & 55\%\tiny{$\pm$3\%} & Uniform Downsampling \\
Random 100k & 100k & 33.7\tiny{$\pm$0.2} & +4.9\tiny{$\pm$0.2} & 59\%\tiny{$\pm$3\%} & Uniform Downsampling \\
ICONS 20k & 20k & 33.6 & +4.8 & 58\% & \citet{wu2024icons} \\
ICONS 50k & 50k & 33.9 & +5.1 & 61\% & \citet{wu2024icons} \\
ARDS 20k & 20k & 33.4 & +4.6 & 55\% & \citet{yang2025data} \\
ARDS 50k & 50k & 33.8 & +5.0 & 60\% & \citet{yang2025data} \\
\bottomrule
\end{tabular}}
\end{table}

Table~\ref{tab:main2} shows results for \textbf{open-prompt} generalist agents on selecting 20k/50k data from LLaVA 665k instruction-tuning dataset for training LLaVA-1.5-7B models. The open-prompt agents achieve general improvements over random baselines and reach or exceed evaluated human baselines across data sizes.

\begin{table}[h]
\centering
\small
\caption{\small{\textbf{Open-prompt} generalist agents on selecting 20k/50k data from LLaVA 665k instruction-tuning dataset for training LLaVA-1.5-7B models. Scores are averaged over 8
benchmarks; ``\% of full gain'' is the fraction of the
base$\to$full-data fine-tuning improvement recovered. \textcolor{Green}{Green}
deltas are over the random baseline at the same budget. Best per
budget in \textbf{bold}.}}\vspace{-0.5em}
\label{tab:main2}
\resizebox{0.75\textwidth}{!}{
\begin{tabular}{l|lll}
\toprule
\textbf{Setting} & \textbf{Avg. score} & \textbf{$\Delta$ over base} & \textbf{\% of full gain} \\
\midrule
Base model (no fine-tuning) & 28.8 & 0.0 & 0\% \\
Full-data fine-tuning (665k) & 37.1 & +8.3 & 100\% \\
\midrule
\multicolumn{4}{l}{\textit{20k budget}} \\
Random (baseline) & 33.0\tiny{$\pm$0.2} & +4.2 & 51\% \\
ICONS \citep{wu2024icons} & 33.6 & +4.8 & 58\% \\
ARDS \citep{yang2025data} & 33.4 & +4.6 & 55\% \\
Agent -- Claude Code (Opus 4.7), 10 iter. & \textbf{33.8 \textcolor{Green}{(+0.8)}} & +5.0 & \textbf{60\% \textcolor{Green}{(+9\%)}} \\
Agent -- Codex (GPT-5.4), 10 iter. & 33.7 \textcolor{Green}{(+0.7)} & +4.9 & 59\% \textcolor{Green}{(+8\%)} \\
\midrule
\multicolumn{4}{l}{\textit{50k budget}} \\
Random (baseline) & 33.4\tiny{$\pm$0.2} & +4.6 & 55\% \\
ICONS \citep{wu2024icons} & 33.9 & +5.3 & 61\% \\
ARDS \citep{yang2025data} & 33.8 \textcolor{Green}{(+0.4)} & +5.0 & 60\% \textcolor{Green}{(+5\%)}\\
Agent -- Codex (GPT-5.4), 10 iter. & \textbf{34.0 \textcolor{Green}{(+0.6)}} & +5.2 & \textbf{63\%\textcolor{Green}{(+8\%)}} \\
\midrule
Random 100k (reference) & 33.7\tiny{$\pm$0.2} & +4.9 & 59\% \\
\bottomrule
\end{tabular}}
\end{table}

Table~\ref{tab:ablation-task} shows results for ablation studies with \textbf{different data tasks (different models/datasets)} on multimodal instruction-tuning data-selection. The open-prompt agents achieve general improvements over random baselines across the tasks.

\begin{table}[h]
\centering
\small
\caption{\small{Results for ablation studies with \textbf{different data tasks (different models/datasets)} on multimodal instruction-tuning data-selection. Scores averaged over 8 benchmarks. \textcolor{Green}{Green} numbers show improvement over the random 10k baseline. Best outcomes are marked in \textbf{bold}. The open-prompt agent (with Claude Code) achieves general improvements over random baselines across the tasks.}}\vspace{-0.5em}
\label{tab:ablation-task}
\resizebox{0.65\textwidth}{!}{
\begin{tabular}{l|l|l|l}
\toprule
\textbf{Setting} & \textbf{Data size} & \textbf{Avg. score} & \textbf{Notes} \\
\midrule
\multicolumn{4}{l}{\textbf{Qwen2-VL-2B + LLaVA-665K dataset}}  \\
\textit{Random 10k (baseline)}& \textit{10k} & \textit{48.1}  & \textit{Uniform Downsampling}\\
Agent (open-prompt)  & 10k & \textbf{48.5  \textcolor{Green}{(+0.4)}} &  Claude Code (Opus 4.7), 10 iter.\\
\midrule
\multicolumn{4}{l}{\textbf{Qwen2.5-VL-3B + LLaVA-665K dataset}}\\
\textit{Random 10k (baseline)}& \textit{10k} & \textit{63.1}  & \textit{Uniform Downsampling}\\
Agent (open-prompt)  & 10k & \textbf{63.5  \textcolor{Green}{(+0.4)}} &  Claude Code (Opus 4.7), 10 iter.\\
\midrule
\multicolumn{4}{l}{\textbf{LLaVA-1.5-7B + Vision-Flan dataset}}\\
\textit{Random 10k (baseline)}& \textit{10k} & \textit{31.8}  & \textit{Uniform Downsampling}\\
Agent (open-prompt)  & 10k & \textbf{32.2  \textcolor{Green}{(+0.4)}} &  Claude Code (Opus 4.7), 10 iter.\\
\bottomrule
\end{tabular}}
\end{table}

\subsection{Example Agent Traces and Trajectory Annotations}
Examples of trace annotations for autonomous data curation with Codex (open-prompt) are provided in Table~\ref{tab:codex-base10}.
\begin{table*}[h]
\centering
\caption{\small{Trace annotations for autonomous data curation with Codex (open-prompt): each row is one commit produced by the agent. ``Keep'' rows improved the composite accuracy; ``discard'' rows were rolled back. The four right-most columns rate each iteration on whether it introduces a new policy family, is grounded in evidence, is effective (improves accuracy), or is shallow/trivial. {\large$\circ$} = yes, {\small\ding{55}} = no. Best accuracy marked in \textbf{bold}.}}
\label{tab:codex-base10}
\resizebox{\textwidth}{!}{%
\begin{tabular}{lcllcccc}
\toprule
\textbf{Commit} & \textbf{Avg.\ score} & \textbf{Status} & \textbf{Description} & \textbf{New Policy} & \textbf{Grounded} & \textbf{Effective} & \textit{\text{Shallow}} \\
\midrule
\texttt{0000000} & 28.93 & keep    & Pre-init baseline (no finetune) & -- & -- & -- & -- \\
\texttt{cf218fe} & 32.53 & keep    & Baseline random 10k & -- & -- & -- & -- \\
\midrule
\texttt{c8cb321} & 32.68 & keep    & Near-random visual, no text\_only & $\circ$ & $\circ$ & $\circ$ & \ding{55} \\
\texttt{28806e1} & 33.24 & keep    & Stratified random reweight & $\circ$ & $\circ$ & $\circ$ & \ding{55} \\
\texttt{778e306} & 32.13 & discard & COCO-rebalance tradeoff & \ding{55} & \ding{55} & \ding{55} & $\circ$ \\
\texttt{d5ac779} & 32.84 & discard & Midpoint mix: stronger MMVet/LLaVA, weak MMBench/Math & \ding{55} & \ding{55} & \ding{55} & $\circ$ \\
\texttt{a19f32c} & 33.01 & discard & Seed sweep: better MMBench/Math, weaker MMVet/OCR & \ding{55} & \ding{55} & \ding{55} & $\circ$ \\
\texttt{ebac17e} & 32.93 & discard & Evenly-spaced within subset quotas & \ding{55} & \ding{55} & \ding{55} & $\circ$ \\
\texttt{bbdf052} & 33.27 & keep    & OCR/textvqa text-rich mixed sampling & $\circ$ & $\circ$ & $\circ$ & \ding{55} \\
\texttt{489dae9} & 32.40 & discard & Seed sweep for OCR text-rich mix & \ding{55} & \ding{55} & \ding{55} & $\circ$ \\
\texttt{e4578e2} & 32.11 & discard & Mixed seeds: main=123, OCR=3407 & \ding{55} & \ding{55} & \ding{55} & $\circ$ \\
\texttt{b88a84f} & \textbf{33.53} & keep    & OCR text-rich ratio 60/40 & \ding{55} & $\circ$ & $\circ$ & \ding{55} \\
\midrule
\multicolumn{2}{r}{\textbf{Execution success rate:} 10/10} & \multicolumn{2}{r}{\textbf{Progression on data:}} & 3/10 & 4/10 & 4/10 & 6/10 \\
\bottomrule
\end{tabular}%
}
\end{table*}
 
Examples of agent trajectory and the corresponding task scores at each iteration for autonomous data curation with Claude Code (open-prompt) are provided in Table~\ref{tab:res-iter-cc-unscaf}.

\begin{table*}[h]
\centering
\caption{\small{Agent trajectory and the corresponding task scores at each iteration for autonomous data curation with Claude Code (open-prompt). Each row is one commit produced by the agent. ``Keep'' rows improved the composite accuracy; ``discard'' rows were rolled back.}}
\label{tab:res-iter-cc-unscaf}
\resizebox{\textwidth}{!}{%
\begin{tabular}{lcccccccccl p{6cm}}
\toprule
\textbf{Commit} & \textbf{Avg. score} & \textbf{MMVet} & \textbf{LLaVABench} & \textbf{OCRBench} & \textbf{HalluBench} & \textbf{MMMU} & \textbf{MathVista} & \textbf{MMStar} & \textbf{MMBench} & \textbf{Status} & \textbf{Description} \\
\midrule
\texttt{0000000} & 28.95 & 17.8 & 26.0 & 240 & 26.8 & 31.80 & 25.7 & 33.8 & 45.60 & keep    & Pre-init baseline (no finetune) \\
\texttt{9327618} & 32.38 & 26.7 & 37.7 & 291 & 29.2 & 31.78 & 24.1 & 30.8 & 49.74 & keep    & Baseline random 10k \\
\midrule
\texttt{d5bf0c3} & 33.06 & 28.0 & 37.3 & 290 & 30.0 & 34.11 & 23.3 & 33.6 & 49.14 & keep    & Balanced subset, OCR boost, no text\_only \\
\texttt{9490f21} & 32.88 & 28.1 & 41.5 & 300 & 27.1 & 33.78 & 22.2 & 32.5 & 47.85 & discard & Balanced subsets + longest asst.\ responses \\
\texttt{117adc1} & 32.06 & 28.1 & 36.9 & 285 & 30.3 & 28.11 & 24.0 & 32.7 & 47.93 & discard & Balanced subsets + text\_only (500) \\
\texttt{8e9876a} & 33.23 & 29.8 & 37.4 & 288 & 32.5 & 30.67 & 23.8 & 34.6 & 48.28 & keep    & Equal allocation 2000 each visual subset \\
\texttt{5647fa1} & 32.56 & 29.0 & 32.8 & 287 & 32.8 & 30.67 & 23.6 & 34.9 & 48.02 & discard & Boost VG/GQA, reduce COCO/OCR \\
\texttt{c6c0c38} & 33.63 & 28.6 & 39.0 & 294 & 30.5 & 34.89 & 23.3 & 33.7 & 49.66 & keep    & Near-equal: COCO=2500, others=1875 \\
\texttt{a32439c} & 32.95 & 29.9 & 34.4 & 288 & 30.4 & 34.56 & 23.5 & 33.2 & 48.93 & discard & COCO=3000, others=1750 (too much COCO) \\
\texttt{72cad7b} & 33.70 & 29.4 & 38.8 & 296 & 31.15 & 34.11 & 23.8 & 33.2 & 49.55 & keep    & OCR-boosted: COCO=2000, OCR/Text=2250 \\
\texttt{85fc5a3} & 33.02 & 28.0 & 35.9 & 296 & 32.64 & 31.33 & 23.3 & 34.9 & 48.45 & discard & heavy OCR+VG: COCO=1500, too little COCO hurts \\
\texttt{2c09740} & 33.74 & 30.4 & 38.9 & 296 & 33.87 & 30.00 & 24.7 & 33.9 & 48.54 & keep    & GQA+textvqa boost: COCO=2000, GQA=2250, textvqa=2250 \\
\bottomrule
\end{tabular}%
}
\end{table*}

Examples of agent trajectory and the corresponding task scores at each iteration for autonomous data curation with Claude Code (Light Scaffold I, Data strategies) are provided in Table~\ref{tab:cc-light-1}.
\begin{table*}[h]
\centering
\caption{Agent trajectory and the corresponding task scores at each iteration for autonomous data curation with Claude Code (Light Scaffold I, Data strategies). Each row is one commit produced by the agent. ``Keep'' rows improved the composite accuracy; ``discard'' rows were rolled back.}
\label{tab:cc-light-1}
\resizebox{\textwidth}{!}{%
\begin{tabular}{lcccccccccl p{6cm}}
\toprule
\textbf{Commit} & \textbf{Avg. score} & \textbf{MMVet} & \textbf{LLaVABench} & \textbf{OCRBench} & \textbf{HalluBench} & \textbf{MMMU} & \textbf{MathVista} & \textbf{MMStar} & \textbf{MMBench} & \textbf{Status} & \textbf{Description} \\
\midrule
\texttt{0000000} & 28.84 & 17.9 & 25.7 & 238 & 26.7 & 31.40 & 25.9 & 33.8 & 45.50 & keep    & Pre-init baseline (no finetune) \\
\texttt{c3ba68c} & 32.49 & 26.6 & 35.3 & 295 & 27.9 & 35.30 & 22.9 & 32.9 & 49.60 & keep    & Random 10k baseline \\
\midrule
\texttt{3d4d1d5} & 32.42 & 25.7 & 36.7 & 295 & 27.8 & 34.80 & 24.6 & 33.1 & 47.20 & keep    & Subset balanced (3k COCO / 2k OCR / 2k GQA / 1.5k VG / 1.5k TextVQA) \\
\texttt{4295574} & 32.31 & 26.5 & 37.5 & 286 & 28.4 & 34.20 & 23.6 & 33.3 & 46.50 & discard & quality-weighted sampling sqrt(asst\_len) - hurt factual benchmarks \\
\texttt{fcf3d46} & 33.14 & 25.5 & 38.7 & 289 & 30.0 & 35.10 & 24.0 & 33.4 & 49.40 & keep    & COCO task diversity + orig quotas + 15\% floor (best in session) \\
\texttt{c7b7c12} & 33.06 & 28.2 & 41.5 & 293 & 27.8 & 33.70 & 22.3 & 32.9 & 49.10 & discard & soft quality pref (longest responses) - verbose COCO bias hurt \\
\texttt{218f5bd} & 32.44 & 29.4 & 36.2 & 291 & 29.8 & 29.30 & 23.5 & 32.1 & 50.20 & discard & OCR-boosted quotas - reducing GQA hurt MMMU badly (29.3) \\
\texttt{fa7b881} & 33.06 & 25.6 & 40.2 & 292 & 28.8 & 35.40 & 22.9 & 32.4 & 49.80 & discard & question-hash stratified sampling - no improvement \\
\texttt{b2f9da3} & 32.42 & 27.2 & 36.7 & 297 & 27.1 & 35.40 & 22.7 & 30.4 & 50.10 & discard & COCO tier-balanced (short/med/long) - no improvement \\
\texttt{4ae030e} & 33.30 & 30.2 & 36.2 & 295 & 29.3 & 35.90 & 23.8 & 31.5 & 49.90 & keep    & 8k samples proportional quotas (fewer=regularization?) \\
\texttt{d5c210c} & 33.21 & 28.2 & 36.6 & 298 & 32.4 & 35.10 & 23.6 & 31.3 & 48.70 & discard & 6k samples - similar to 8k, slightly worse MMBench \\
\texttt{9f08a09} & 33.97 & 29.7 & 40.3 & 303 & 29.5 & 35.60 & 23.6 & 33.0 & 49.80 & keep    & 8k COCO-heavy (3k coco, 1k vg, 1.5k ocr, 1.5k gqa, 1k textvqa) - NEW BEST \\
\bottomrule
\end{tabular}%
}
\end{table*}

Examples of agent trajectory and the corresponding task scores at each iteration for autonomous data curation with Claude Code (Light Scaffold II, Research papers) are provided in Table~\ref{tab:cc-light-2}.
\begin{table*}[h]
\centering
\caption{Agent trajectory and the corresponding task scores at each iteration for autonomous data curation with Claude Code (Light Scaffold II, Research papers). Each row is one commit produced by the agent. ``Keep'' rows improved the composite accuracy; ``discard'' rows were rolled back.}
\label{tab:cc-light-2}
\resizebox{\textwidth}{!}{%
\begin{tabular}{lcccccccccl p{5cm}}
\toprule
\textbf{Commit} & \textbf{Avg. score} & \textbf{MMVet} & \textbf{LLaVABench} & \textbf{OCRBench} & \textbf{HalluBench} & \textbf{MMMU} & \textbf{MathVista} & \textbf{MMStar} & \textbf{MMBench} & \textbf{Status} & \textbf{Description} \\
\midrule
\texttt{0000000} & 28.95 & 17.8 & 26.0 & 240 & 26.8 & 31.80 & 25.7 & 33.8 & 45.60 & keep    & Pre-init baseline (no finetune) \\
\texttt{f9b769f} & 32.19 & 26.6 & 35.9 & 292 & 28.9 & 32.60 & 24.0 & 30.3 & 50.10 & keep    & Baseline random 10k \\
\midrule
\texttt{72df18e} & 32.18 & 28.0 & 33.6 & 285 & 31.7 & 31.20 & 23.9 & 31.1 & 49.30 & discard & balanced subset sampling with OCR boost \\
\texttt{b8e2f93} & 32.09 & 28.5 & 38.7 & 292 & 26.9 & 31.60 & 23.4 & 31.4 & 47.00 & discard & quality-based: longest responses per subset \\
\texttt{3dd1aee} & 33.29 & 29.1 & 32.4 & 290 & 32.7 & 34.80 & 24.2 & 34.6 & 49.50 & keep    & minimal COCO no text\_only boost visual diversity \\
\texttt{d47b299} & 32.81 & 27.9 & 34.3 & 290 & 35.1 & 29.30 & 24.2 & 35.2 & 47.50 & discard & near-zero COCO 500 boost VG 3k OCR 2.5k \\
\texttt{199733f} & 32.77 & 29.0 & 34.0 & 286 & 33.2 & 33.40 & 23.3 & 32.3 & 48.40 & discard & text\_only 500 back, ocr\_vqa 1500 \\
\texttt{55275b8} & 33.41 & 27.4 & 35.6 & 286 & 33.0 & 35.10 & 24.1 & 35.3 & 48.20 & keep    & 7k samples same proportions \\
\texttt{6565487} & 32.76 & 27.8 & 37.0 & 297 & 31.0 & 33.80 & 22.7 & 33.6 & 46.60 & discard & 7k quality-filtered remove shortest 25\% per subset \\
\texttt{5941975} & 33.72 & 30.1 & 37.9 & 283 & 32.6 & 35.30 & 24.2 & 33.5 & 47.80 & keep    & 8k samples winning proportions scaled up \\
\texttt{1e7b23b} & 32.67 & 28.3 & 37.3 & 279 & 33.9 & 26.70 & 24.2 & 34.1 & 48.90 & discard & 9k samples winning proportions \\
\texttt{679ba47} & 32.20 & 29.7 & 33.8 & 279 & 33.5 & 26.70 & 23.7 & 34.5 & 47.90 & discard & 8k boost OCR/TextVQA reduce VG/GQA \\
\bottomrule
\end{tabular}%
}
\end{table*}
 
Examples of agent trajectory and the corresponding task scores at each iteration for autonomous data curation with Claude Code (Heavy Scaffold I, Self-research) are provided in Table~\ref{tab:cc-heavy-1a}.
\begin{table*}[h]
\centering
\caption{Agent trajectory and the corresponding task scores at each iteration for autonomous data curation with Claude Code (Heavy Scaffold I, Self-research). Each row is one commit produced by the agent. ``Keep'' rows improved the composite accuracy; ``discard'' rows were rolled back.}
\label{tab:cc-heavy-1a}
\resizebox{\textwidth}{!}{%
\begin{tabular}{lcccccccccl p{6cm}}
\toprule
\textbf{Commit} & \textbf{Avg. score} & \textbf{MMVet} & \textbf{LLaVABench} & \textbf{OCRBench} & \textbf{HalluBench} & \textbf{MMMU} & \textbf{MathVista} & \textbf{MMStar} & \textbf{MMBench} & \textbf{Status} & \textbf{Description} \\
\midrule
\texttt{0000000} & 28.95 & 17.8 & 26.0 & 240 & 26.8 & 31.80 & 25.7 & 33.8 & 45.60 & keep    & Pre-init baseline (no finetune) \\
\texttt{f9b769f} & 32.19 & 26.6 & 35.9 & 292 & 28.9 & 32.60 & 24.0 & 30.3 & 50.10 & keep    & Baseline random 10k \\
\midrule
\texttt{8fcb60b} & 32.61 & 26.1 & 39.9 & 285 & 28.2 & 33.80 & 23.0 & 31.9 & 49.50 & keep    & drop text\_only, stratified visual 10k \\
\texttt{b3f1382} & 33.12 & 26.4 & 39.0 & 299 & 29.1 & 34.70 & 23.9 & 31.8 & 50.30 & keep    & drop text\_only+vg, stratified 4-subset 10k \\
\texttt{0bf0abb} & 32.51 & 27.7 & 36.4 & 290 & 28.5 & 34.80 & 23.6 & 30.9 & 49.20 & discard & 2x OCR weight oversampling \\
\texttt{58a5268} & 31.47 & 24.8 & 39.3 & 290 & 27.1 & 26.00 & 24.4 & 32.3 & 49.00 & discard & multi-turn coco priority (all 3+ turns) \\
\texttt{424754d} & 32.59 & 25.7 & 37.1 & 295 & 27.5 & 34.60 & 23.7 & 32.3 & 50.30 & discard & reduce GQA to 500, extra to coco \\
\texttt{ead8bac} & 32.32 & 24.4 & 37.1 & 293 & 26.5 & 35.20 & 24.7 & 31.1 & 50.20 & discard & proportional SEED=7 (variance test) \\
\texttt{0dc4697} & 32.65 & 27.2 & 39.9 & 288 & 28.0 & 34.10 & 23.0 & 32.4 & 47.80 & discard & COCO short-bias reduction (25\% short / 75\% non-short) \\
\texttt{933dce7} & 32.24 & 26.9 & 40.4 & 287 & 27.4 & 28.70 & 22.9 & 32.1 & 50.80 & discard & OCR rebalance (ocr\_vqa-400, textvqa+400) \\
\texttt{5da628e} & 31.75 & 28.2 & 39.0 & 288 & 26.6 & 26.00 & 22.8 & 31.9 & 50.70 & discard & GQA diversity filter (4+ unique question types) \\
\texttt{d2755be} & 33.32 & 26.3 & 41.8 & 289 & 26.8 & 35.60 & 23.7 & 31.9 & 51.60 & keep    & ocr\_vqa-200 to coco for visual diversity \\
\bottomrule
\end{tabular}%
}
\end{table*}

Examples of agent logs on "Failure Mode" and "Next Skill Candidate" at each iteration for autonomous data curation corresponding to the trajectory in Table~\ref{tab:cc-heavy-1a} are provided in Table~\ref{tab:cc-heavy-1b}.

\begin{table*}[h]
\centering
\caption{Agent logs on "Failure Mode" and "Next Skill Candidate" at each iteration for autonomous data curation corresponding to the trajectory in Table~\ref{tab:cc-heavy-1a}. Each row is one commit produced by the agent. ``Keep'' rows improved the composite accuracy; ``discard'' rows were rolled back.}
\label{tab:cc-heavy-1b}
\resizebox{\textwidth}{!}{%
\begin{tabular}{lcl p{3cm} p{8cm} p{6.5cm}}
\toprule
\textbf{Commit} & \textbf{Avg. score} & \textbf{Status} & \textbf{Description} & \textbf{Failure Mode} & \textbf{Next Skill Candidate} \\
\midrule
\texttt{0000000} & 28.95 & keep    & Pre-init baseline (no finetune) & weak on OCR and math; hallucination issues & none (random baseline to run) \\
\texttt{f9b769f} & 32.19 & keep    & Baseline random 10k & weak math/logical reasoning; OCR improved over pre-init & (first skill to explore) \\
\midrule
\texttt{8fcb60b} & 32.61 & keep    & drop text\_only, stratified visual 10k & LLaVABench +4.0 but OCR regressed $-$7; removing text\_only helped conversation/detail but lost some OCR signal; gqa short answers may be diluting useful signal & which coco samples resemble OCRBench/MathVista test items? \\
\texttt{b3f1382} & 33.12 & keep    & drop text\_only+vg, stratified 4-subset 10k & OCR recovered vs iter1; MMMU +1.9 over baseline; MathVista still weak at 23.9; gqa short-answer training may not transfer to complex reasoning & can we improve detailed reasoning by biasing coco towards longer, multi-step responses? \\
\texttt{0bf0abb} & 32.51 & discard & 2x OCR weight oversampling & OCR oversampling didn't improve OCRBench (290 vs 299); reduced coco budget hurt LLaVABench/MMBench; coco diversity > OCR-specific volume & within coco, do detailed multi-turn conversations transfer better than short-answer samples? \\
\texttt{58a5268} & 31.47 & discard & multi-turn coco priority (all 3+ turns) & MMMU crashed 34.7$\to$26.0; removing 1-turn MC samples destroys MC-format performance; diversity within coco matters more than multi-turn bias & what makes iter2's proportional split work well? try different seeds to measure selection variance \\
\texttt{424754d} & 32.59 & discard & reduce GQA to 500, extra to coco & HallusionBench dropped $-$1.6; GQA yes/no format helps hallucination discrimination more than expected; proportional balance is hard to beat & can we improve by keeping proportional but using diversity sampling within subsets? \\
\texttt{ead8bac} & 32.32 & discard & proportional SEED=7 (variance test) & seed variance confirmed: $-$0.008 vs SEED=42; MMMU/MathVista slightly better but MMVet/LLaVA/Hall worse; seed affects benchmarks differently & try SEED=100 to test whether higher seeds produce better draws \\
\texttt{0dc4697} & 32.65 & discard & COCO short-bias reduction (25\% short / 75\% non-short) & short COCO samples help OCR/MMBench more than hurt reasoning; OCRBench $-$11 and MMBench $-$2.5; MathVista unchanged & do OCR\_VQA and short COCO samples share training signal? try removing redundant ones \\
\texttt{933dce7} & 32.24 & discard & OCR rebalance (ocr\_vqa$-$400, textvqa+400) & OCR\_VQA yes/no+factual format helps MMMU MC format (28.7 vs 34.7); TextVQA captioning does not help OCRBench & why does OCR\_VQA help MMMU? is it the yes/no MC-format training? \\
\texttt{5da628e} & 31.75 & discard & GQA diversity filter (4+ unique question types) & MMMU needs repetitive yes/no format, not diverse Q types; diversity dilutes MC-format signal; high-diversity GQA has more turns which changes training dynamics & can image-level dedup within COCO improve visual diversity without changing format? \\
\texttt{d2755be} & 33.32 & keep    & ocr\_vqa$-$200 to coco for visual diversity & OCRBench $-$10 as expected; HallusionBench $-$2.3 unexpected; confirms MMMU crash was from TextVQA not OCR\_VQA; COCO diversity helps LLaVA/MMBench & can we recover HallusionBench by tweaking GQA or adding hallucination-specific COCO samples? \\
\bottomrule
\end{tabular}%
}
\end{table*}

Examples of agent trajectory and the corresponding task scores at each iteration for autonomous data curation with Claude Code (Heavy Scaffold II, Adapt papers) are provided in Table~\ref{tab:cc-heavy-2a}.
\begin{table*}[h]
\centering
\caption{Agent trajectory and the corresponding task scores at each iteration for autonomous data curation with Claude Code (Heavy Scaffold II, Adapt Papers). Each row is one commit produced by the agent. ``Keep'' rows improved the composite accuracy; ``discard'' rows were rolled back.}
\label{tab:cc-heavy-2a}
\resizebox{\textwidth}{!}{%
\begin{tabular}{lcccccccccl p{7cm}}
\toprule
\textbf{Commit} & \textbf{Avg. score} & \textbf{MMVet} & \textbf{LLaVABench} & \textbf{OCRBench} & \textbf{HalluBench} & \textbf{MMMU} & \textbf{MathVista} & \textbf{MMStar} & \textbf{MMBench} & \textbf{Status} & \textbf{Description} \\
\midrule
\texttt{0000000} & 28.95 & 17.8 & 26.0 & 240 & 26.8 & 31.80 & 25.7 & 33.8 & 45.60 & keep    & Pre-init baseline (no finetune) \\
\texttt{f9b769f} & 32.19 & 26.6 & 35.9 & 292 & 28.9 & 32.60 & 24.0 & 30.3 & 50.10 & keep    & Baseline random 10k \\
\midrule
\texttt{6fe2c47} & 31.88 & 27.0 & 37.6 & 284 & 28.9 & 30.40 & 23.7 & 30.9 & 48.10 & discard & D4 hash-dedup + BoW k-means diversification \\
\texttt{03b05c0} & 31.93 & 29.6 & 38.2 & 290 & 29.1 & 28.70 & 22.8 & 32.1 & 46.00 & discard & DEITA LLM-judge quality+complexity scoring on 30k pool \\
\texttt{2505edd} & 31.53 & 25.7 & 36.7 & 284 & 27.2 & 29.30 & 24.1 & 31.5 & 49.30 & discard & Self-Instruct MinHash dedup (Jaccard>0.7) + random sample \\
\texttt{692fab3} & 32.02 & 27.1 & 36.0 & 287 & 29.1 & 27.30 & 24.6 & 32.3 & 51.10 & discard & Cambrian visual-centric: images only + exact dedup + random \\
\texttt{58a6fab} & 31.49 & 27.0 & 38.5 & 282 & 29.4 & 24.70 & 24.8 & 30.1 & 49.30 & discard & SemDeDup TF-IDF question clustering, random per cluster \\
\texttt{9602a05} & 33.71 & 27.7 & 32.7 & 279 & 31.1 & 36.40 & 25.1 & 32.6 & 56.20 & keep    & EL2N proxy: base model forward-pass loss, top-10k highest-loss from 50k pool \\
\texttt{6177e53} & 31.99 & 27.1 & 37.9 & 285 & 27.6 & 30.80 & 23.9 & 31.3 & 48.80 & discard & Moderate-difficulty band (p25--p90): random 10k from loss band \\
\texttt{9dc1d88} & 33.70 & 26.1 & 35.2 & 299 & 31.3 & 35.10 & 25.2 & 33.1 & 53.80 & discard & D4 diversification on EL2N top-20k: TF-IDF k-means 20 clusters, 500 highest-loss per cluster \\
\texttt{4694fbf} & 34.86 & 29.1 & 40.4 & 290 & 29.2 & 36.30 & 25.7 & 32.1 & 57.10 & keep    & EL2N top-loss with p95 assistant-loss noise filter \\
\texttt{7bb6675} & 33.72 & 24.5 & 37.9 & 284 & 30.0 & 33.20 & 23.7 & 34.1 & 57.80 & discard & Coreset: 100k pool, p95 noise filter, TF-IDF k-means 20 clusters, top-loss per cluster \\
\bottomrule
\end{tabular}%
}
\end{table*}

Examples of agent logs on "Failure Mode" and "Next Skill Candidate" at each iteration for autonomous data curation corresponding to the trajectory in Table~\ref{tab:cc-heavy-2a} are provided in Table~\ref{tab:cc-heavy-2b}.

\begin{table*}[h]
\centering
\caption{Agent logs on "Failure Mode" and "Next Skill Candidate" at each iteration for autonomous data curation corresponding to the trajectory in Table~\ref{tab:cc-heavy-2a}. Each row is one commit produced by the agent. ``Keep'' rows improved the composite accuracy; ``discard'' rows were rolled back.}
\label{tab:cc-heavy-2b}
\resizebox{\textwidth}{!}{%
\begin{tabular}{lcl p{5cm} p{8.5cm} p{5.5cm}}
\toprule
\textbf{Commit} & \textbf{Avg. score} & \textbf{Status} & \textbf{Description} & \textbf{Failure Mode} & \textbf{Next Skill Candidate} \\
\midrule
\texttt{0000000} & 28.95 & keep    & Pre-init baseline (no finetune) & weak on OCR and math; hallucination issues & none (random baseline to run) \\
\texttt{f9b769f} & 32.19 & keep    & Baseline random 10k & weak math/logical reasoning; OCR improved over pre-init & (first skill to explore) \\
\midrule
\texttt{6fe2c47} & 31.88 & discard & D4 hash-dedup + BoW k-means diversification & BoW text clustering too crude for semantic diversity; OCR and MMMU regressed & DEITA: proxy quality+complexity scoring \\
\texttt{03b05c0} & 31.93 & discard & DEITA LLM-judge quality+complexity scoring on 30k pool & quality scoring biased toward verbose responses; MMMU and MMBench regressed ($-$3.9/$-$4.1) & Beyond Neural Scaling Laws: moderate-difficulty selection \\
\texttt{2505edd} & 31.53 & discard & Self-Instruct MinHash dedup (Jaccard>0.7) + random sample & aggressive dedup removed 20\% of data including valid question variations; hurt OCR/MMMU/Hallusion & Cambrian: vision-centric data balance \\
\texttt{692fab3} & 32.02 & discard & Cambrian visual-centric: images only + exact dedup + random & improved 5/8 benchmarks (MMStar+2.0, MMBench+1.0) but MMMU regressed $-$5.3 (lost text\_only language transfer) & SemDeDup: TF-IDF question-based clustering + Cambrian visual-only \\
\texttt{58a6fab} & 31.49 & discard & SemDeDup TF-IDF question clustering, random per cluster & equal-per-cluster sampling destroyed natural distribution; MMMU $-$7.9 (academic content severely undersampled) & EL2N: base model forward-pass loss as data quality proxy \\
\texttt{9602a05} & 33.71 & keep    & EL2N proxy: base model forward-pass loss, top-10k highest-loss from 50k pool & LLaVABench and OCR regressed ($-$3.2/$-$13) suggesting highest-loss includes noisy/corrupt samples that hurt generation quality & Dataset Cartography: filter ambiguous vs hard-but-wrong to remove noisy high-loss samples \\
\texttt{6177e53} & 31.99 & discard & Moderate-difficulty band (p25--p90): random 10k from loss band & moderate band is too similar to random; lost reasoning signal (MMMU $-$5.6, MMBench $-$7.4) from high-loss samples while LLaVABench recovered (+5.2) & LIMA: fewer but higher-quality; EL2N top-5k to test if concentration beats dilution \\
\texttt{9dc1d88} & 33.70 & discard & D4 diversification on EL2N top-20k: TF-IDF k-means 20 clusters, 500 highest-loss per cluster & near-matched EL2N (0.3370 vs 0.3371); OCR recovered (+20) and LLaVABench (+2.5) but MMVet/MMBench regressed slightly & ShareGPT4V: response quality scoring; decompose loss into user/assistant tokens \\
\texttt{4694fbf} & 34.86 & keep    & EL2N top-loss with p95 assistant-loss noise filter & HallusionBench slightly regressed ($-$1.9) suggesting some filtered samples were actually hallucination-correcting examples & Active Learning: coreset selection on EL2N-scored pool for maximum diversity coverage \\
\texttt{7bb6675} & 33.72 & discard & Coreset: 100k pool, p95 noise filter, TF-IDF k-means 20 clusters, top-loss per cluster & equal per-cluster allocation (500/cluster) forced low-quality samples from small clusters; MMVet $-$4.6, MMMU $-$3.1 vs best & TRAK: scale up pool to 100k with pure p95 filter (no diversity) \\
\bottomrule
\end{tabular}%
}
\end{table*}

\clearpage
\section{Extended Discussion}
\label{app:discussion}

\subsection{From autonomous execution to agent-assisted research}

Our results support a sharp distinction between two modes of agent progress. In \textbf{autonomous execution}, the human supplies a relatively concrete objective and the agent performs steps with clear local verification (fixing code, running training, producing a valid submission). In \textbf{agent-assisted research}, the objective is underspecified, the policy space is discontinuous, and evaluation is sparse and noisy. Training-data curation belongs to the second category: there is no oracle that tells the agent whether a data strategy is scientifically meaningful before an expensive training run.

This distinction reframes the research agenda. Execution tasks may continue to benefit primarily from stronger models, better tools, and longer contexts. Data research additionally requires \emph{better scaffolds}--representations of the data-policy space, mechanisms for grounding hypotheses, protocols for comparing policies under noise, and interfaces that turn prior research into executable agent actions.

\subsection{Scaffolds as research artifacts}

A scaffold is not merely a prompt; it is an experimental-design object that determines what kinds of reasoning and action the agent can perform. In our experiments, a weak scaffold produces plans while a strong scaffold produces valid data policies--so the same backbone can look like a different system depending on its scaffold. Future papers should therefore report scaffold details with the same rigor as model details: a score obtained under a detailed pipeline with examples is not directly comparable to a score obtained under an open-ended instruction.

\subsection{The autonomy--performance trade-off}

Heavy scaffolds can improve performance while reducing autonomy, and this is a feature to measure rather than a flaw to avoid. A fully open task asks whether the agent can \emph{discover} the method; a heavily scaffolded task asks whether it can \emph{execute and adapt} the method. Both are useful, but they answer different questions, and the field will be better served by reporting a scaffold-strength ladder--open instruction, broad prompt, skill list, executable skill card, scripted pipeline--rather than a single number.

\section{Broader Impacts and Considerations}
\label{app:broader}

\subsection{Broader Societal Impacts}

\bench{} and the accompanying findings sit at the intersection of two
trajectories with mixed societal implications: the increasing capability of
generalist coding agents and the increasing centrality of training-data
curation to model behavior. We discuss the most direct positive and negative
impacts we anticipate.

\paragraph{Potential positive impacts.}
First, automating training-data curation can substantially reduce the human
labor and compute required to build effective models. Our strongest observed $10$k subset recovers up to $71\%$ of the
full-data fine-tuning gain on $665$k examples, suggesting that
agent-curated subsets can lower the barrier to fine-tuning for practitioners with limited compute, including academic
laboratories, public-sector users, and groups in low-resource settings who
cannot afford full-pool training. Second, the framework makes data decisions
more legible: trajectory diagnostics record \emph{why} a given dataset was
selected, which supports auditability of model training pipelines and is
useful for downstream accountability. Third, by open-sourcing \bench{} and the
scaffold ladder, we aim to make agent-driven data research a shared public
artifact rather than a proprietary capability concentrated in a few labs.

\paragraph{Potential negative impacts.}
The same capabilities admit several concerning failure modes.
\emph{(i) Reduced human oversight of training data.} If agent-curated subsets
are deployed without review, harmful biases, unsafe content, or
underrepresented groups in the candidate pool may be silently amplified or
removed by opaque scoring functions. 
\emph{(ii) Acceleration of capability development for misuse.} A general
recipe for data optimization can in principle be applied to data pools whose
downstream models are designed for harmful tasks (e.g., disinformation,
targeted persuasion, surveillance-related vision tasks). The scaffolds we
study--particularly the paper-adaptation scaffold that turns published methods
into executable agent actions--reduce the engineering friction for any party
attempting such optimization, not only beneficial ones.
\emph{(iii) Concentration of advantage.} The strongest results we report
require capable backbone agents and non-trivial compute for repeated
training/evaluation. If only well-resourced actors can run the heavy-scaffold
loop effectively, agentic data curation could widen rather than narrow the gap
between large industrial labs and the broader community.

\subsection{Considerations and Safeguards}
\paragraph{Mitigations we recommend.}
We encourage three concrete practices when building on this work.
First, treat agent-curated datasets as candidate artifacts subject to human
or programmatic review for representation, safety, and license compliance,
rather than as final training inputs. Second, report scaffold strength and
trajectory-level diagnostics alongside outcome scores, so that improvements
can be attributed to the agent's actual research process rather than to
opaque tuning. Third, retain contamination checks and held-out evaluations
when operating agents that can read evaluation logs; \bench{} ships such
checks and we recommend they remain enabled in derivative pipelines.

\subsection{Released Artifacts and Residual Risk}

The artifacts we release are a benchmark harness, scaffolds (prompt
specifications), trace logs, and curated dataset \emph{manifests} (i.e.,
indices into public datasets). We do not release new pretrained models or
new raw multimodal datasets. We have nonetheless taken several precautions.

\paragraph{Source data.}
All candidate pools are existing public datasets--LLaVA-665K
\citep{liu2024improved} and Vision-Flan~\citep{xu2024vision}--used under
their original licenses. Manifests select subsets of these pools rather than
redistributing images or text, which preserves the upstream license posture
and avoids creating a new venue for problematic content.

\paragraph{Traceability and legibility.}
Every iteration's curation script, manifest, audit result, training output,
evaluation scores, and run notes are persisted under a commit hash
(P4 in Section~\ref{sec:bench}). This supports post-hoc inspection of agent
behavior and makes it possible to identify, retract, or correct individual
iterations if downstream issues are discovered.

\paragraph{Scope of the released artifact.}
We judge the misuse risk of releasing \bench{} itself to be modest: it does
not provide novel offensive capabilities, it operates on already-public
datasets, and its primary contribution is methodological--an evaluation
framework and scaffolds for studying agent behavior. We will release the
benchmark with documentation that explicitly describes the contamination
audit, the assumption that downstream users will retain it, and the
recommendation that agent-curated datasets be reviewed before deployment.

\subsection{Declaration of LLM Use}
LLMs are central components of this work and are declared
accordingly.

\paragraph{LLMs as the object of study.}
The agents we evaluate--Claude Code with Opus 4.7 and Sonnet 4.6, and Codex
with GPT-5.4 and GPT-5.3--are themselves LLM-based systems, and they are
the experimental subject rather than auxiliary tools. All claims about
agent behavior, exploration, and scaffold sensitivity are claims about
these systems running under the protocols described in
Sections~\ref{sec:prototype},\ref{sec:scaffolds}.

\paragraph{LLMs as tools inside the pipeline.}
Three additional LLM-driven components are part of the methodology rather
than the writing process. \emph{(i) Trace annotation.} The four trajectories
labels in Table~\ref{tab:rubric} (new policy, grounded, effective, shallow)
are produced with LLM assistance using Claude Opus 4.7, following the
rubric in Section~\ref{sec:bench_diagnostics}. We treat these labels as diagnostic annotations rather than ground-truth
scientific claims, and we provide rubrics and trace examples to support
auditing. \emph{(ii) Skill-card compilation.}
The paper-derived skill library used by light scaffold II and heavy scaffold
II is generated by an automated Claude Code pipeline against a corpus of
data-curation papers, following the template in
Appendix~\ref{app:prompts}. \emph{(iii) Rewriting tool and judge.}
The rewriting experiments call Qwen3.5-2B/9B and Gemma3-4B as rewriting
backbones, and judge-based VLMEvalKit metrics use a separate Qwen3.5-27B
endpoint (\Cref{tab:judge-server-config}). All of these are documented with
prompts, model versions, and configurations sufficient for reproduction.

\paragraph{LLMs for writing assistance.}
Beyond the components above, LLMs were used only in standard writing-support
roles (grammar, phrasing, and LaTeX formatting suggestions). All scientific
claims, experimental designs, analyses, and conclusions are the authors'.
This use of LLMs does not affect the methodology, results, or originality of
the work, and per the LLM policy is reported here for transparency
rather than as a methodological dependency.

\section{Prompts and Skill-Card Templates}\label{app:prompts}

Additional full prompts and example skill cards are provided in the code repository.

\begin{center}
\begin{tcolorbox}[enhanced, breakable, width=0.95\linewidth,
  title=\textbf{Self-Research Iteration: Shift 200 OCR\_VQA to COCO},
  colback=lightgray!10!white, colframe=black]
\small

\# Research: Shift 200 OCR\_VQA to COCO (not TextVQA)\\

\#\# Observation\\

Three prior experiments changed training data composition. All crashed MMMU:\\
-- 933dce7 (ocr\_vqa $-$400, textvqa +400): MMMU 34.7 $\to$ 26.0\\
-- 5da628e (GQA diversity filter): MMMU 34.7 $\to$ 26.0\\
-- 0dc4697 (COCO short-bias): MMMU 34.7 $\to$ 34.1 (minor)\\

The 933dce7 crash was attributed to OCR\_VQA reduction, but it also doubled TextVQA. TextVQA is 100\% captioning format (``Provide a one-sentence caption''), which is radically different from MMMU's MC format. The MMMU crash may have been from TextVQA's captioning format conflicting with MC-format capability, not from OCR\_VQA reduction.\\

Supporting evidence: 0bf0abb increased OCR weight (oversampled OCR\_VQA) and MMMU was 34.8 -- essentially unchanged. This suggests OCR\_VQA count changes don't affect MMMU much, and 933dce7's crash was from TextVQA doubling.\\

OCR\_VQA content: 39\% yes/no, 21\% ``Who wrote this book?'', 21\% ``What is the title?'' -- highly formulaic book metadata extraction. Of the 4 retained subsets, OCR\_VQA is the most formulaic and least diverse.\\
 
\#\# Hypothesis\\
Shifting 200 samples from OCR\_VQA to COCO (NOT TextVQA) will:\\
1. Keep MMMU stable (no TextVQA change, OCR\_VQA change is small)\\
2. Improve visual diversity (200 more unique COCO images)\\
3. Potentially improve MMVet/LLaVABench (+0.3--0.5 from more detailed COCO content)\\
4. Minor OCRBench impact ($-$3 to $-$5 from slightly less OCR training)\\

Expected net: +0.001 to +0.003 accuracy\\

\#\# Minimal Change\\

After computing proportional allocations:\\
-- ocr\_vqa: 1486 $\to$ 1286 ($-$200)\\
-- coco: 6765 $\to$ 6965 (+200)\\
-- gqa: 1340 (unchanged)\\
-- textvqa: 408 (unchanged)

\normalsize
\end{tcolorbox}
 \end{center}

\begin{center}
\begin{tcolorbox}[enhanced, breakable, width=0.95\linewidth,
  title=\textbf{Instructions}  for automated compilation of skill cards compiled from research papers,
  colback=lightgray!10!white, colframe=black]
\small
\#\# Characteristics of the skills\\

-- Four top-level buckets, each a stage of the VLM data pipeline: data-acquisition, data-curation, data-selection, data-synthesis\\
-- Each skill is one folder containing exactly one file: `SKILL.md` with a title that combines the paper's short title and source.\\
-- 1 paper -> 1 skill. No multi-paper aggregations; no extra assets (no code, no figures, no PDFs).\\

**Per-file structure (rigidly templated)**\\
Every `SKILL.md` uses the same H2 sections in the same order:\\

1. `\# \{Paper title\}`\\
2. **One-line decision** — when to use / when not to (single sentence each).\\
3. **Skill metadata** — `skill type`, `paper kind` (e.g. `operational-method`), `actionability` (high/med/low), `evidence quality` (`full\_paper`).\\
4. **Goal** — one paragraph operational restatement.\\
5. **Problem signature** — modality, data state, scale regime, model requirement.\\
6. **Use when** / **Do not use when** — bullet pairs.\\
7. **Required inputs** / **Optional inputs** / **Outputs** — bolded named slots with one-line descriptions.\\
8. **Assumptions and prerequisites**.\\
9. **Procedure** — numbered steps, each with `Action: … / Why: … / Note: See paper for details.`\\
10. **Parameters to set** — Role / How to set / Default-or-range / Effect.\\
11. **Validation checks**, **Failure modes**.\\
12. **Adaptation notes for VLM training** — always present, ties the paper back to VLM data work even when the source paper is not VLM-specific.\\
13. **Implementation notes**.\\
14. **Evidence from the paper** — 3–5 bullets with the load-bearing claims.\\
15. **Source paper** — title, year, venue, paper ID, URL, arXiv ID.\\

**Tone / philosophy**\\
-- Decision-first ("Use this skill when…" / "Avoid it when…") rather than paper summary.\\
-- Procedures are skeletons — the *shape* of the method, not full reproduction. Each step ends with "See paper for details."\\
-- Knowledge-cards optimized for a downstream agent to retrieve and act on, not for human reading. They look like cookbook recipe cards keyed by paper.
\normalsize
\end{tcolorbox}
 \end{center}

\begin{center}
\begin{tcolorbox}[enhanced, width=0.95\linewidth,
  title=\textbf{Skill cards} compiled from research papers on \textit{data curation vision-language models (VLMs)},
  colback=lightgray!10!white, colframe=black]
\small
\#\# Characteristics of the skills\\

-- Four top-level buckets, each a stage of the VLM data pipeline: data-acquisition, data-curation, data-selection, data-synthesis\\
-- Each skill is one folder containing exactly one file: `SKILL.md` with a title that combines the paper's short title and source.\\
-- 1 paper -> 1 skill. No multi-paper aggregations; no extra assets (no code, no figures, no PDFs).\\

**Per-file structure (rigidly templated)**\\

**Tone / philosophy**\\
-- Decision-first ("Use this skill when…" / "Avoid it when…") rather than paper summary.\\
-- Procedures are skeletons — the *shape* of the method, not full reproduction. Each step ends with "See paper for details."\\
-- Knowledge-cards optimized for a downstream agent to retrieve and act on, not for human reading. They look like cookbook recipe cards keyed by paper.
\normalsize
\end{tcolorbox}
\end{center}

\begin{center}
\begin{tcolorbox}[enhanced, width=0.95\linewidth,
  title=\textbf{Light-scaffolding I: }General Data Strategists,
  colback=lightgray!10!white, colframe=black]
\small
\#\# Curation strategies to explore\\

Ideas (not limited to these):\\

-- **Subset balancing**: Analyze source\_subset distribution and adjust proportions.\\
-- **Quality filtering**: Score by text quality (length, complexity, vocabulary diversity).\\
-- **Diversity sampling**: Cover diverse image types and question categories.\\
-- **Task-type targeting**: Prioritize samples matching eval benchmarks (OCR, visual reasoning, open-ended QA).\\
-- **Deduplication**: Remove near-duplicates to make room for diverse samples.\\
-- **Difficulty scoring**: Balanced mix of easy/medium/hard.\\
-- **Conversation structure**: Prioritize multi-turn or specific conversation patterns.\\
-- **Image analysis**: Use image metadata (resolution, content type) to guide selection.
\normalsize
\end{tcolorbox}
\end{center}
\begin{center}
\begin{tcolorbox}[enhanced, breakable, width=0.95\linewidth,
  title=\textbf{Heavy-scaffolding: }Self-research (rewriting),
  colback=lightgray!10!white, colframe=black]
\small
\begin{itemize}[leftmargin=0.5em, itemsep=1pt]
    \item \textit{Mandatory LLM rewriting after baselines}: Every post-baseline iteration must apply the MLLM provided as a tool to rewrite at least part of the training set (>20\% samples recommended, >50\% encouraged). Selection-only or programmatic-only iterations are explicitly banned. 

\item \textit{Research-first mandate with eval-row grounding}: Similar to "self-research scaffolds", this version requires writing a research plan at each iteration, requiring the observation to cite specific failure rows from raw evaluation outputs (question, model output, gold answer), not just aggregate scores. A research plan that only references aggregated scores is explicitly declared insufficient.

\item \textit{Rewrite-recipe research questions}: a set of examples is provided as starting points: answer-length mismatches, format gaps, malformed samples, turn restructuring, ambiguous questions. These frame the kind of hypotheses the agent should form, distinct from previous selection-oriented thinking.

\item \textit{Rewrite verification step}: After curation/rewriting but before training, the agent must eyeball 5–10 rewritten samples to confirm the transform actually did what the hypothesis predicted. If the rewrite is generic or didn't change what was expected, the agent must fix the recipe before burning a training run. The result logs gains a `rewrite recipe` column describing the specific LLM rewrite applied (target slice + transform).

\item \textit{Failure analysis section}: Requires reading raw evaluation files after scoring — sampling 5–15 wrong rows per benchmark, categorizing failures (OCR error, format violation, hallucinated option, etc.), and cross-referencing with training data to find the linkage between what the model learned and where it fails. 

\item \textit{Revised no-contamination rule}: An explicit "No data contamination" rule forbids using evaluation questions or answers (from eval logs) to construct training examples — not as copies, paraphrases, or rewrite targets mirroring specific eval items. Patterns are fair game; specific items are not. 
\end{itemize}
\normalsize
\end{tcolorbox} 
\end{center}

\begin{center}
\begin{tcolorbox}[enhanced, width=0.95\linewidth,
  title=\textbf{Adapt papers:} EL2N top-loss with p95 assistant-loss noise filter (removes top 5\% noisiest),
  colback=lightgray!10!white, colframe=black]
\small
Data curation: ShareGPT4V-inspired response quality scoring.\\

\textbf{Skill:} sharegpt4v-improving-large-multi-modal-models-with-better-captions-arxiv-2311-12793v2\\

\textbf{Adaptation:} ShareGPT4V improves VLM training by ensuring high-quality captions/responses. 
We adapt this by decomposing the base model's forward-pass loss into total loss (EL2N
informativeness) and assistant-only loss (response coherence quality). Select samples 
with high total loss (informative) but NOT extreme assistant-only loss (coherent responses). 
This filters out high-loss samples whose difficulty comes from noisy/incoherent responses 
rather than genuinely complex visual understanding.
\normalsize
\end{tcolorbox}
\end{center}
\end{appendices}
\end{document}